%% file: main.tex
\documentclass[10pt]{article}

\usepackage[letterpaper,left=0.978in,right=0.978in,top=0.80in,bottom=0.92in,footskip=0.45in]{geometry}

\usepackage[T1]{fontenc}
\usepackage[utf8]{inputenc}
\input{math_commands.tex}

\usepackage{amsthm}
\usepackage{amsmath}
\usepackage{graphicx}
\usepackage[table]{xcolor}
\usepackage{booktabs}
\usepackage{multirow}
\usepackage{tabularx}
\usepackage{caption}
\usepackage{float}
\usepackage{placeins}
\usepackage{needspace}
\usepackage{wrapfig}
\usepackage{microtype}
\usepackage[scaled=0.98]{helvet}
\usepackage[most]{tcolorbox}
\usepackage{titlesec}
\usepackage[numbers,sort]{natbib}
\usepackage{xurl}

\definecolor{ink}{HTML}{1C2B33}
\definecolor{linkviolet}{HTML}{5D2CD6}
\definecolor{urlmagenta}{HTML}{EC008C}
\definecolor{boxbg}{gray}{0.95}

\usepackage{hyperref}
\hypersetup{colorlinks=true,linkcolor=linkviolet,citecolor=linkviolet,urlcolor=urlmagenta}

\renewcommand{\topfraction}{0.92}

\renewcommand{\textfraction}{0.07}
\renewcommand{\floatpagefraction}{0.75}

\makeatletter
\def\@listi{\leftmargin\leftmargini
            \labelwidth\leftmargini\advance\labelwidth-\labelsep
            \topsep 2\p@ \@plus 1\p@ \@minus 1\p@
            \parsep \z@
            \itemsep 3\p@ \@plus 1\p@ \@minus 1\p@}
\makeatother

\titleformat{\section}
  {\color{black}\sffamily\bfseries\fontsize{12.5}{15}\selectfont}
  {\thesection}{0.85em}{}
\titlespacing*{\section}{0pt}{17pt plus 3pt minus 2pt}{9pt plus 1pt}
\titleformat{\subsection}
  {\color{black}\sffamily\bfseries\fontsize{10.5}{13}\selectfont}
  {\thesubsection}{0.85em}{}
\titlespacing*{\subsection}{0pt}{12pt plus 2pt minus 2pt}{6pt plus 1pt}
\titleformat{\subsubsection}
  {\color{black}\sffamily\bfseries\fontsize{10}{12}\selectfont}
  {\thesubsubsection}{0.85em}{}
\titlespacing*{\subsubsection}{0pt}{10pt plus 2pt minus 2pt}{5pt plus 1pt}

\theoremstyle{plain}

\theoremstyle{definition}

\theoremstyle{remark}

\title{Read, Allocate, Route: Role-Coupled Tokenization for Pixel-Space Generative Transformers}
\author{}
\date{}

\begin{document}
\color{ink}

\begin{tcolorbox}[colback=boxbg,colframe=boxbg,sharp corners,boxrule=0pt,
                  left=16pt,right=16pt,top=15pt,bottom=14pt,
                  before skip=0pt,after skip=14pt]
\color{ink}
{\color{black}\sffamily\bfseries\fontsize{17}{20.5}\selectfont
Read, Allocate, Route: Role-Coupled Tokenization for Pixel-Space Generative Transformers\par}
\vspace{0.85em}

{\sffamily\bfseries\fontsize{8.5}{11}\selectfont\color{black}
Jiongfan Yang\textsuperscript{1,3},\quad
Weichen Fan\textsuperscript{2},\quad
Haiwen Diao\textsuperscript{2},\quad
Zhongang Cai\textsuperscript{3},\quad
Ziwei Liu\textsuperscript{2,$\dagger$}\par}
\vspace{0.35em}

{\fontsize{10}{12}\selectfont
\textsuperscript{1}Peking University,\quad
\textsuperscript{2}Nanyang Technological University,\quad
\textsuperscript{3}SenseTime,\quad
\textsuperscript{$\dagger$}Corresponding author\par}
\vspace{0.8em}

Pixel-space generative Transformers typically use one token per image patch, so capturing finer detail requires more tokens and makes global attention increasingly expensive. We observe that finer detail does not necessarily require more global communication: low-frequency structure benefits from global exchange, while high-frequency detail can be modeled locally. Based on this insight, we introduce SpecSlot, a pixel-space rectified-flow Transformer that decouples output detail from the length of the global attention sequence. SpecSlot expands each spatial anchor into multiple slots, each assigned a disjoint band of DCT coefficients; this separates anchor density from spectral subdivision. Only the lowest-frequency slot from each anchor participates in global attention, while all slots interact locally. This allows the model to increase spectral detail without enlarging the global sequence. A spatial head further adds a local correction to the spectral prediction, while the generative state, target, and loss remain in pixel space. On ImageNet-256, SpecSlot achieves a comparable final FID to a FLOPs-matched JiT control while converging faster. On ImageNet-512, it improves FID by 2.84 over the control with 3.9\% fewer FLOPs, and comes within only 0.69 FID of JiT-B/16 at 38.2\% lower cost. On text-to-image generation, SpecSlot surpasses JiT-B/8 on DPG-Bench by 1.06 at 40.6\% lower forward cost.

\vspace{0.55em}
{\hspace*{13pt}\sffamily\bfseries\fontsize{7.5}{10}\selectfont\color{black}\raisebox{-0.25\height}{\tikz[x=1em,y=1em,line width=0.4pt]{\draw (0,0) circle (0.5);\draw (0,0) ellipse (0.23 and 0.5);\draw (-0.5,0)--(0.5,0);}}\;Project Page:\quad{\fontsize{9}{11}\selectfont\url{https://yangjiongfan.site/specslot/}}\par}
\vspace{0.2em}
{\hspace*{13pt}\sffamily\bfseries\fontsize{7.5}{10}\selectfont\color{black}\raisebox{-0.25\height}{\includegraphics[height=1.05em]{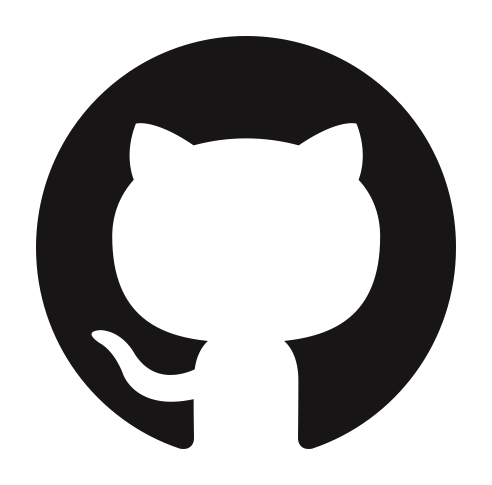}}\;GitHub:\quad{\fontsize{9}{11}\selectfont\url{https://github.com/JiongfanYang/SpecSlot}}\par}
\vspace{0.2em}
{\hspace*{13pt}\sffamily\bfseries\fontsize{7.5}{10}\selectfont\color{black}\raisebox{-0.25\height}{\includegraphics[height=1.05em]{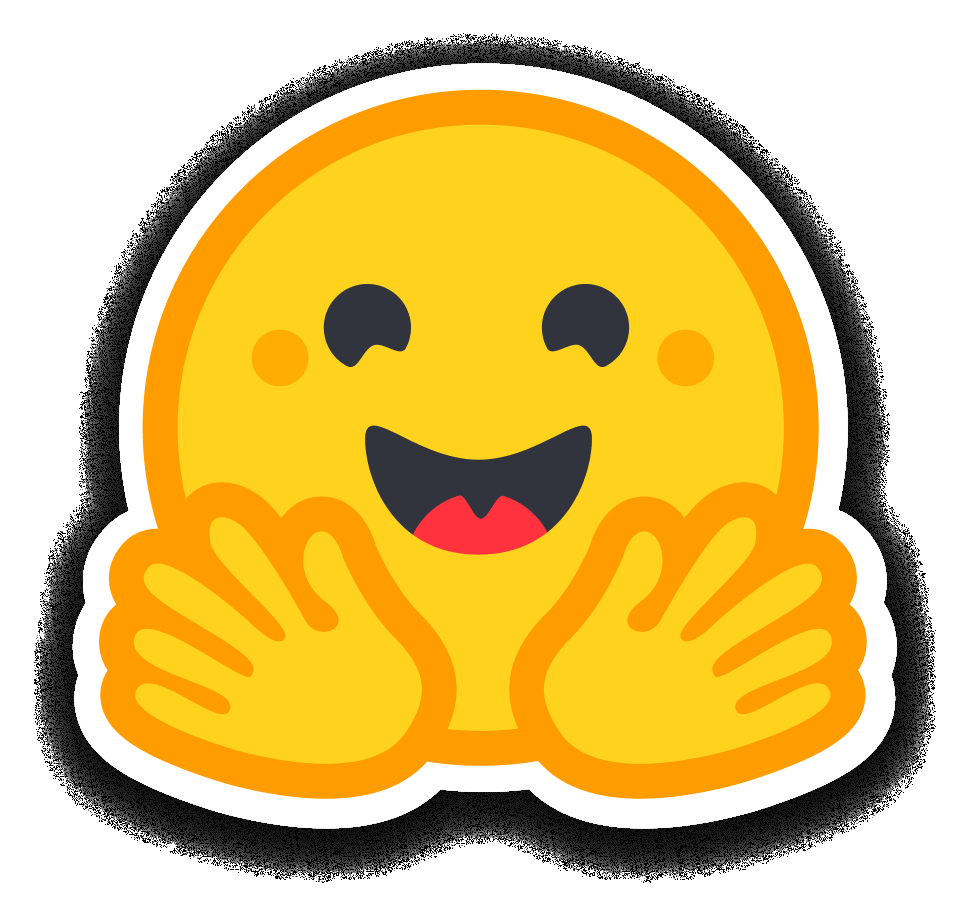}}\;Hugging Face:\quad{\fontsize{9}{11}\selectfont\url{https://huggingface.co/Gokottaw434/SpecSlot}}\par}
\end{tcolorbox}

\begin{center}
\vspace{2pt}
\includegraphics[width=\textwidth]{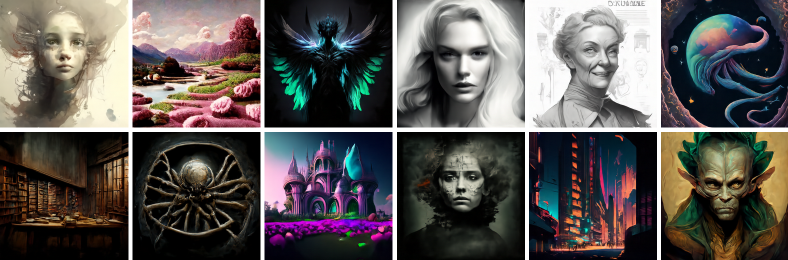}
\captionof{figure}{Text-to-image samples generated by SpecSlot. More samples with prompts are shown in Appendix Figure~\ref{fig:t2i-qualitative-appendix}.}
\label{fig:t2i-qualitative-front}
\end{center}
\vspace{4pt}

\input{sections/introduction}
\input{sections/related_work}
\input{sections/method}
\input{sections/experiments}
\input{sections/ablation}
\input{sections/mechanistic}
\input{sections/conclusion}

\bibliography{references}
\bibliographystyle{unsrtnat}

\appendix
\renewcommand{\topfraction}{0.95}
\renewcommand{\textfraction}{0.05}
\renewcommand{\floatpagefraction}{0.75}
\input{sections/appendix}

\end{document}

%% file: math_commands.tex
\usepackage{amsmath,amsfonts,bm}

\def\eqref#1{equation~\ref{#1}}

\def\1{\bm{1}}

\DeclareMathAlphabet{\mathsfit}{\encodingdefault}{\sfdefault}{m}{sl}
\SetMathAlphabet{\mathsfit}{bold}{\encodingdefault}{\sfdefault}{bx}{n}



%% file: sections/introduction.tex
\section{Introduction}
\label{sec:intro}

Pixel-space generative models learn directly from noise to pixels, eliminating the separately trained autoencoder and its compression bottleneck. This simplicity, however, comes with a computational cost: the full-resolution image must be represented as Transformer tokens. Most pixel-space Transformers use one token per spatial patch, so the patch grid simultaneously determines what the model reads, what each token predicts, and how many tokens participate in global attention. As a result, finer image detail directly increases attention cost: on a $256\times256$ image, reducing the patch size from $16\times16$ to $8\times8$ increases the token count from 256 to 1{,}024, making the quadratic term of dense attention $16\times$ larger.

Existing methods address this cost from different directions, but largely retain the coupling between output detail and token communication. JiT routes every patch token through the global sequence~\citep{li2026basicsletdenoisinggenerative}. Multiscale systems move detail to extra stages or detailer heads~\citep{Yu_2026_CVPR,chen2026diptamingdiffusionmodels,he2026hyperdithyperconnectedtransformershighfidelity}, and efficient-attention methods shorten the sequence with generic patterns---windows, selected global tokens, or learned sparsity~\citep{9710580,beltagy2020longformerlongdocumenttransformer,jaegle2021perceivergeneralperceptioniterative,zhou2026trainableloglinearsparseattention}. Spectral Forcing brings in frequency structure, but only to filter the noisy input; its tokens remain spatial patches~\citep{fan2026signalhidenoisespectral}. Across the pixel-space generators discussed above, output detail remains tied to a spatial patch grid, even when communication is compressed or restricted.

Our key observation is that finer detail does not necessarily require proportionally more global communication: in the frequency domain, low-frequency structure carries broad image organization and benefits from long-range exchange, while fine detail is largely local. A token carrying broad structure therefore stands to gain from exchanging with the whole image, while a token carrying fine detail mostly does not. Yet in existing designs, what a token communicates is fixed by where it sits, not by what it produces. Can a token's output role also determine how it should communicate?

\begin{figure}[t]
    \centering
    \includegraphics[width=\textwidth]{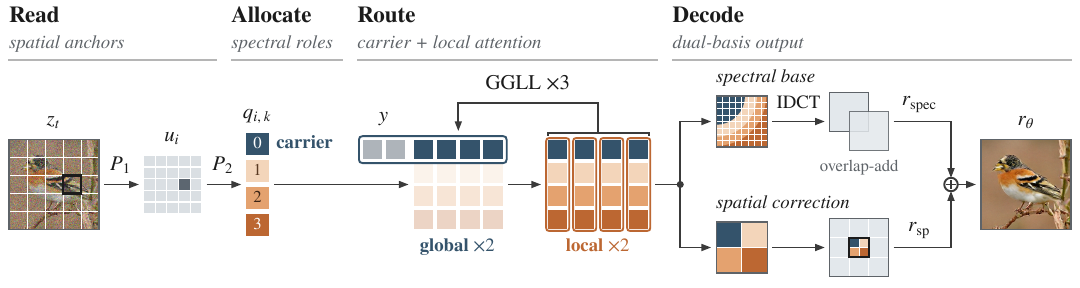}
\caption{SpecSlot data flow. Each anchor expands into $K$ slots. G blocks process lowest-radial slot-0 carriers across anchors with context states; L blocks mix all slots within each anchor. The final slot states decode a lapped-DCT spectral base and an anchor-aligned spatial correction, summed in pixel space.}
    \label{fig:sd-architecture}
\end{figure}

SpecSlot builds on this idea by making output roles explicit and using them to control communication. It organizes computation around three operations: \emph{Read}, \emph{Allocate}, and \emph{Route}. Read forms a grid of spatial anchors that collect input evidence. Allocate expands each anchor into $K$ indexed slots, with each slot assigned a disjoint radial group of DCT coefficients. Route then uses these spectral roles to decide communication: only the slot assigned the lowest-frequency group from each anchor enters the global stage, while all slots interact locally within the anchor. This separates fine-grained output representation from the length of the global sequence. A spatial head further reuses the same slot states to add a local correction to the spectral prediction, without introducing additional tokens. Figure~\ref{fig:sd-architecture} illustrates this design.

This representation exposes two separate allocation variables. $W$ controls the density of spatial anchors, while $K$ controls the spectral subdivision at each anchor. The total number of slots grows with $WK$, but the global image sequence contains only $W$ carriers. A fixed token budget can therefore be allocated either to denser spatial anchors or to finer spectral roles without forcing both to increase together. This turns tokenization from a single patch-size choice into an explicit quality--compute allocation problem.

Our contributions are:
\begin{itemize}
    \item We introduce an indexed $W\times K$ representation in which spatial anchors collect input evidence and slots receive disjoint spectral coefficient groups. The split exposes anchor density and coefficient subdivision as separate allocation variables, and the same slot states also decode a local pixel correction.
    \item We define carrier-routed global--global--local--local (GGLL) attention from the slot index: one lowest-radial carrier per anchor enters global blocks, and all slots enter anchor-local blocks, so the global sequence contains $W$ carriers rather than $WK$ slots.
    \item On ImageNet-256 and ImageNet-512, SpecSlot provides two measured quality--compute operating points between JiT's low-cost and high-cost settings. At ImageNet-512, W256/K4 lowers FID by 2.84 over our FLOPs-matched JiT control (JiT-CM) with 3.9\% fewer GFLOPs, and trails JiT-B/16 by 0.69 FID at 38.2\% lower cost. Larger output windows cover more pixels while the global sequence remains fixed; the larger improvement over JiT-CM is consistent with this within-anchor allocation. On text-to-image generation, SpecSlot likewise surpasses JiT-B/8 on DPG-Bench by 1.06 at 40.6\% lower forward cost.
\end{itemize}

%% file: sections/related_work.tex
\section{Related Work}
\label{sec:related}

\paragraph{Diffusion and pixel-space generation.}
Diffusion and score-based models learn to reverse a gradual noising process and provide the iterative foundation of modern image synthesis~\citep{3495724.3496298,song2021scorebasedgenerativemodelingstochastic,NEURIPS2021_49ad23d1}. Rectified flow turns the same idea into a straight noise-to-data path~\citep{liu2022flowstraightfastlearning}. Latent diffusion moved this process into the latent space of a separately trained autoencoder, and improved autoencoders continue to strengthen this line~\citep{9578911,9878449,zheng2025diffusiontransformersrepresentationautoencoders}. DiT, U-ViT, and SiT showed that plain Transformers scale reliably in this setting~\citep{10377858,bao2023worthwordsvitbackbone,ma2024sitexploringflowdiffusionbased}. Pixel-space generation removes the autoencoder, but then asks the model to process the full-resolution state. This line has followed several routes: Simple Diffusion and Hourglass DiT reduce the state internally, cascaded models and PixelFlow distribute generation across resolutions, and JiT shows that a direct large-patch Transformer can remain a strong pixel-space baseline~\citep{hoogeboom2023simplediffusionendtoenddiffusion,3692070.3692447,3586589.3586636,chen2025pixelflowpixelspacegenerativemodels,li2026basicsletdenoisinggenerative}. Detail recovery is then assigned to extra components: PixelDiT adds a pixel-level DiT, DiP adds a detailer head, HyperDiT connects coarse and fine flows, and PixNerd decodes patch tokens with a neural field~\citep{Yu_2026_CVPR,chen2026diptamingdiffusionmodels,he2026hyperdithyperconnectedtransformershighfidelity,wang2025pixnerdpixelneuralfield}. These designs differ in where the compute goes, but share one premise: a token is a spatial patch, so finer detail and wider communication stay tied to the same spatial grid. SpecSlot complements this line with a single-backbone allocation: indexed slots carry fixed output and communication roles, so detail can be subdivided without making every output token global.

\paragraph{Frequency structure and transform-domain generation.}
Frequency offers a complementary view of this problem. It has entered generative modeling through the state or forward process, through generation order or conditioning, and through the training objective or analysis. In the first line, inverse heat dissipation, wavelet models, sparse transform codes, and DCTdiff change the representation or process used for generation~\citep{rissanen2023generative,NEURIPS2022_03474669,10204642,nash2021generatingimagessparserepresentations,ning2025dctdiff}. In the second, VAR and NFIG organize generation by scale or frequency, while Spectral Forcing and Frequency-Forcing use frequency-related conditioning or guidance; Latent Forcing instead reorders the trajectory with a semantic auxiliary stream~\citep{tian2024visual,huang2025nfigmultiscaleautoregressiveimage,fan2026signalhidenoisespectral,baade2026latent,du2026frequencyforcing}. Focal frequency loss reweights spectral errors, and Fourier-space analyses characterize diffusion without changing the denoiser~\citep{9710849,falck2025fourierspaceperspectivediffusion}. Recent architectural methods make frequency roles part of the denoiser itself: DiMSUM alternates spatial and wavelet-subband processing, while DeCo assigns low and high frequencies to a DiT and a separate pixel decoder~\citep{phung2025dimsumdiffusionmamba,ma2026decofrequencydecoupledpixeldiffusion}. Together these works show several ways for frequency to shape representation or generation, but the cited lines do not use fixed spectral groups as both per-token output ownership and communication eligibility within one pixel-space backbone. SpecSlot keeps the state, target, and loss in pixel space; fixed, disjoint DCT groups give slots spectral roles, and the role index also determines which slots communicate globally.

\paragraph{Visual tokenization and allocation.}
Vision Transformers tokenize an image into a spatial patch grid~\citep{dosovitskiy2021an}, and NaViT extends this idea to variable resolutions~\citep{dehghani2023patch}. This line reduces sequence work by changing how many spatial tokens are retained or processed at each step: ToMe merges similar tokens, SparseDiT prunes them, DyDiT++ reschedules computation across denoising time and space, variable-length tokenization learns a global merge, and FlexiDiT spends fewer tokens at easier steps~\citep{10208676,chang2025sparsedit,zhao2026dyditdiffusiontransformerstimestep,lee2026variablelengthtokenizationlearnableglobal,anagnostidis2025flexiditdiffusiontransformereasily}. The spatial patch remains the allocation unit, so detail and communication stay coupled to the same grid. SpecSlot adds a second, spectral axis: each of $W$ anchors owns $K$ slots with disjoint coefficient groups, making the fixed budget $N{=}WK$ an explicit split between anchor density and spectral subdivision.

\paragraph{Global-local and sparse communication.}
The communication literature provides the other half of the problem by restricting how tokens exchange information. Swin uses local windows with shifted boundaries, Longformer combines local windows with a few global tokens, and Perceiver routes information through a small latent bottleneck~\citep{9710580,beltagy2020longformerlongdocumenttransformer,jaegle2021perceivergeneralperceptioniterative}. In pixel-space generation, RINs exchange information through learned interface states, while trainable sparse attention and PixSGR learn or select interaction patterns from the data~\citep{jabri2023scalableadaptivecomputationiterative,zhou2026trainableloglinearsparseattention,you2026advancedpixeldiffusionmodel}. These approaches provide useful local or global topologies, but do not tie communication eligibility to a fixed spectral output role. SpecSlot fixes the graph from the allocation: one lowest-frequency slot per anchor is the global carrier, and all slots interact locally, so cross-anchor communication is spent on the role that also carries the low-frequency scaffold.

%% file: sections/method.tex
\section{Methodology}
\label{sec:method}

SpecSlot organizes pixel-space generation around one idea: tokenization as allocation---assigning each token a fixed output role and dividing a fixed token budget between anchor density and spectral subdivision. Global attention is then spent only on the role that needs it; the rectified-flow state, target, and loss remain in pixel space.

\subsection{Shared Pixel-Space Rectified-Flow Objective}
\label{sec:pixel-flow}

Following the JiT controls and objectives~\citep{li2026basicsletdenoisinggenerative,liu2022flowstraightfastlearning}, we mix an image $x\in[-1,1]^{3\times H\times H}$ with noise $\epsilon\sim\mathcal N(0,I)$ on the linear path
\begin{equation}
z_t=t x+(1-t)\epsilon,
\label{eq:rf-path}
\end{equation}
where $t=\sigma(g)$ with $g\sim\mathcal N(-0.8,0.8^2)$ and $y$ is the class label. The network returns an image-shaped prediction $r_\theta(z_t,t,y)$, converted to a velocity by
\begin{equation}
v_\theta(z_t,t,y)=
\frac{r_\theta(z_t,t,y)-z_t}{\max(1-t,\tau)},
\qquad \tau=0.05,
\label{eq:x-to-v}
\end{equation}
where the clamp avoids the singularity at $t=1$, and training minimizes the velocity MSE
\begin{equation}
\mathcal L_{\mathrm{RF}}=
\mathbb E_{x,\epsilon,t,y}
\left[\frac{1}{3H^2}
\left\|v_\theta(z_t,t,y)-(x-\epsilon)\right\|_2^2\right].
\label{eq:rf-loss}
\end{equation}

\subsection{SpecSlot Architecture}
\label{sec:factorized-tokenizer}

\begin{figure}[t]
    \centering
    \includegraphics[width=0.78\textwidth]{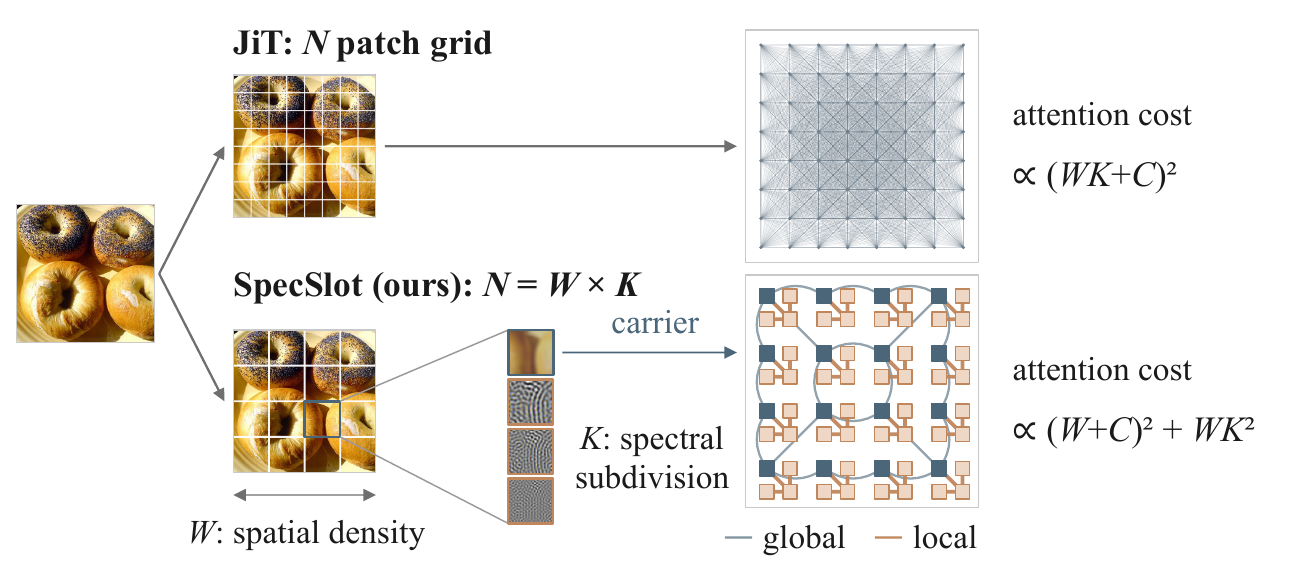}
    \caption{SpecSlot allocation. A patch grid spends all $N$ tokens in one global sequence; splitting $N{=}WK$ into $W$ spatial anchors and $K$ spectral slots keeps global attention at $W$ carriers.}
    \label{fig:motivation}
\end{figure}

Why is allocation needed? In a patch-grid model, adding finer detail means shrinking the patch, and the global sequence grows with it; the two cannot be adjusted separately. Prior designs keep the spatial patch as the only allocation unit. SpecSlot instead introduces the missing variables, forming $W$ spatial anchors and expanding each anchor into $K$ indexed slots (Figure~\ref{fig:motivation}):
\begin{equation}
\begin{aligned}
N&=WK,\qquad L=2s,\\
W&=(H/s)^2,\qquad E=\frac{L^2}{K}=\frac{4H^2}{N}.
\end{aligned}
\label{eq:allocation}
\end{equation}

The budget splits into $W$ anchors that read the input and $K$ slots per anchor that divide the output; each slot owns an equal share $E$ of the coefficients per channel, with $s$ the input cell side and $L$ the output window side. The two variables act separately: $W$ sets how densely the input is read, and $K$ how finely the output is divided. At fixed $N$, anchor density trades directly against spectral subdivision, and output detail can grow with $K$ while the global sequence stays at $W$ carriers. Equal $N$ does not imply equal cost, since communication and projection scale differently with the split.

Three operations implement this allocation, as shown in Figure~\ref{fig:sd-architecture}: Read collects each anchor's input evidence, Allocate assigns each slot a fixed output role, and Route lets the role decide communication.

\subsubsection{Read: Spatial Anchors and Slot Expansion}
\label{sec:spatial-evidence}

The first question is where input evidence comes from; SpecSlot reads it on a grid of anchors, independent of how the output is divided. Let $\mathcal P_i$ extract the $i$-th non-overlapping $s\times s$ cell of $z_t$. A shared projection maps each cell to an anchor vector,
\begin{equation}
u_i=P_1\operatorname{vec}(\mathcal P_i z_t)\in\mathbb R^B,
\qquad P_1\in\mathbb R^{B\times3s^2}.
\label{eq:anchor-projection}
\end{equation}

This projection is a patch embedding whose width $B$ is free, because the output count no longer depends on it. A zero-initialized residual depthwise $3\times3$ convolution over the anchor grid---the input mixer---lets neighboring cells exchange evidence before it is split into roles. A shared $1\times1$ projection expands each mixed anchor vector into $K$ hidden states,
\begin{equation}
 [q_{i,0};\ldots;q_{i,K-1}]
 =\operatorname{reshape}\!\left(P_2\widetilde u_i+b_2\right),
 \qquad q_{i,k}\in\mathbb R^d,
 \label{eq:slot-expansion}
\end{equation}
where $P_2\in\mathbb R^{Kd\times B}$; a two-dimensional anchor position plus a learned slot position are added. One anchor vector seeds all $K$ slots; the roles come from the slot positions, output heads, and routing, not from the input. All $WK$ slots are evaluated at every flow time.

\subsubsection{Allocate: Radial Coefficients and Pixel Synthesis}
\label{sec:spectral-responsibility}

\paragraph{Radial coefficient assignment.}
The second question is who produces what. SpecSlot assigns each slot a fixed share of the output in a DCT basis. Low radial frequencies carry broad structure and high frequencies fine detail, and natural-image spectra decay smoothly along this order~\citep{fan2026signalhidenoisespectral}; a radial split therefore sorts the roles from global to local without any learned assignment. For an output window of side $L=2s$, the DCT-II indices $(u,v)$ are ordered by $\sqrt{u^2+v^2}$ and split into $K$ equal groups, requiring $K\mid L^2$; slot $k$ of each anchor owns group $k$, that is, $E=L^2/K$ coefficients per channel. The disjoint groups cover the grid: every coefficient is predicted exactly once. A shared affine head predicts the owned coefficients from the slot's final hidden state $h^{\mathrm{out}}_{i,k}$:
\begin{equation}
\widehat c_{i,k}=P_{\mathrm{spec}}h^{\mathrm{out}}_{i,k}+b_{\mathrm{spec}}
\in\mathbb R^{3E}.
\label{eq:spectral-head}
\end{equation}
Sharing the head is deliberate: the slot index, not per-band parameters, defines ownership.

\paragraph{Lapped synthesis.}
Decoding each window independently would leave neighbors free to disagree on their shared edge, and the mismatch shows as block seams, most visibly early in training. SpecSlot decodes by the synthesis half of a 50\%-lapped DCT frame instead: inverse DCT, a square-root periodic Hann window, and normalized overlap-add. Neighboring windows share half their pixels, and normalized overlap-add mitigates cross-window discontinuities.
\begin{equation}
r_{\mathrm{spec}}=
\operatorname{OLA}\!\left(\left\{\operatorname{IDCT}(\widehat c_i)\right\}_{i=1}^{W}\right),
\label{eq:lapped-synthesis}
\end{equation}
where each window is weighted before overlap-add. Let $g$ be the separable square-root periodic Hann window, $\mathcal D$ the orthonormal two-dimensional DCT-II, $R_i^\top$ the placement of window $i$ at stride $s=L/2$, and $Q$ the crop to the image region after reflection padding by $s/2$. The normalized synthesis is
\begin{equation}
\begin{aligned}
r_{\mathrm{spec}}&=Q\left[\Lambda^\dagger\odot\sum_i R_i^\top
\left(g\odot\mathcal D^\top\widehat c_i\right)\right],\\
\Lambda&=\sum_i R_i^\top(g^2).
\end{aligned}
\label{eq:normalized-lapped-synthesis}
\end{equation}
Here $\Lambda^\dagger$ is the reciprocal on the covered region. For the matched analysis frame, let $P$ reflect-pad the image by $s/2$ pixels and let $R_i$ extract the $i$-th window. Its coefficients are
\begin{equation}
c_i(x)=\mathcal D\left(g\odot R_iPx\right).
\label{eq:lapped-analysis}
\end{equation}
Applying the synthesis above to these coefficients gives $S(\{c_i(x)\})=x$ up to floating-point error: the same window is used in analysis and synthesis, $\Lambda$ normalizes the overlap, and the crop satisfies $QP=I$.

\paragraph{Anchor-aligned spatial correction.}
The same slot states can also be read out directly in pixel space. A second head adds an anchor-aligned correction alongside the spectral base: slot $k$ writes tile $k$ of a deterministic row-major $K_h\times K_w$ tiling ($K_hK_w{=}K$) of its anchor cell (four $8\times8$ quadrants for $K{=}4$, $s{=}16$). These projections are slot-specific and zero-initialized, so training starts from a pure spectral model. The correction adds output parameters but no token states or attention edges. Radial ownership applies only to the spectral base; the spatial term is a pixel readout with no frequency assignment. The two terms express the same prediction in different bases, and Section~\ref{sec:hybrid-mechanism} finds their content complementary. The prediction is
\begin{equation}
r_\theta(z_t,t,y)=r_{\mathrm{spec}}+r_{\mathrm{sp}}.
\label{eq:dual-basis-output}
\end{equation}

\subsubsection{Route: Carrier and Local Attention}
\label{sec:routing}

The third question is who needs to communicate globally. Sending every slot through global attention would make the global quadratic cost grow with $K$---reintroducing exactly the coupling between output detail and communication cost that the slot allocation set out to remove. The spectral roles suggest a cheaper split: the lowest radial group holds the broad structure that benefits from cross-anchor exchange, while finer groups are increasingly local; its signal also overtakes the noise earliest in the flow (Section~\ref{sec:routing-mechanism}). SpecSlot therefore routes by role: the owner of the lowest radial group, slot $0$, is the carrier and the only slot entering the global stage; all $K$ slots then mix locally within their anchor. Since global and local blocks alternate, the carrier brings global information into its anchor and takes local detail back into the next global stage.

Let $\mathbf Z\in\mathbb R^{W\times K\times d}$ be the slot tensor and $\mathbf Y\in\mathbb R^{C\times d}$ the $C{=}32$ context states formed from the class embedding. The two routing sets are
\begin{equation}
\mathcal A_G=\operatorname{Concat}(\mathbf Y,\mathbf Z_{:,0}),
\qquad
\mathcal A_L^{(i)}=\mathbf Z_{i,:}.
\label{eq:routing-sets}
\end{equation}
Two consecutive blocks process the global set and two more process each local set; this global-global-local-local (GGLL) schedule repeats three times, giving six global and six local blocks of identical dimensions. Class context appears only in the global group. Non-carrier slots traverse the six local blocks, context states the six global ones, and carriers all twelve. On ImageNet-256 with $s{=}16$ and $K{=}4$, the global sequence holds $W{+}C=288$ tokens instead of $WK{+}C=1{,}056$.

Ignoring projections and elementwise operations, the quadratic terms are
\begin{equation}
\begin{aligned}
\mathcal C_{\mathrm{GGLL}}^{\mathrm{quad}}
&\propto 6(W+C)^2d+6WK^2d,\\
\mathcal C_{\mathrm{full}}^{\mathrm{quad}}
&\propto 12(WK+C)^2d,
\end{aligned}
\label{eq:ggll-complexity}
\end{equation}
where $d$ is the hidden width; the global quadratic term now grows with $W$, not $WK$. Measured GFLOPs include the full execution and decoder, not only these terms.

Because $W$ and $K$ can be varied independently, their effects on quality and compute can be measured separately.

%% file: sections/experiments.tex
\section{Experiments}
\label{sec:experiments}

\subsection{Experimental Setup}
\label{sec:experimental-setup}

\paragraph{Models and training.}
We train class-conditional models on ImageNet-1K~\citep{5206848} and report results at ImageNet-256 and ImageNet-512. Unless stated otherwise, our models use the $W256$ anchor grid of Section~\ref{sec:method} with $B{=}1536$, where $B$ is the input-projection width; we write SpecSlot-K2 and SpecSlot-K4 for the two allocations. ImageNet-256 training lasts 120 epochs at effective batch 512, and the aligned models share the full training recipe.

\paragraph{Evaluation.}
\begin{wrapfigure}{r}{0.5\textwidth}
    \centering
    \includegraphics[width=\linewidth]{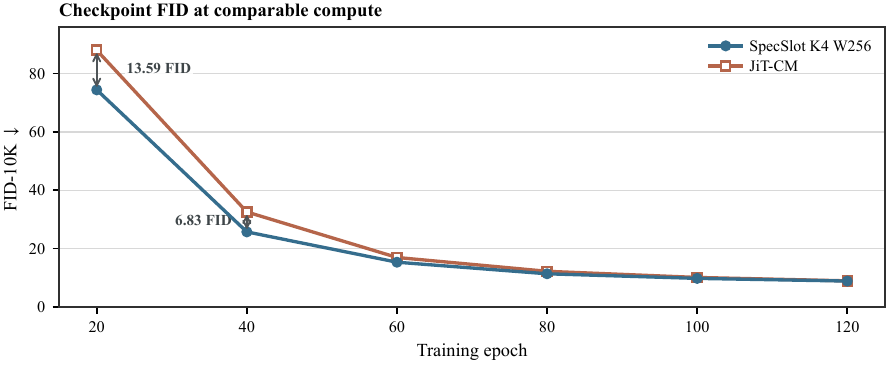}
    \caption{Mean FID-10K under the aligned protocol.}
    \label{fig:training-convergence}
    \vspace{-8pt}
\end{wrapfigure}
We sample with 50 Heun intervals on the common legacy-uniform grid and use EMA1 weights. Reported rows use the per-model best classifier-free guidance (CFG)~\citep{ho2022classifierfreediffusionguidance} from a coarse-to-fine sweep (Appendix~\ref{app:cfg-sweep}); JiT rows use 2.9 instead of 3.0, also the swept optimum for JiT (Table~\ref{tab:cfg-sweep}), matching SF's protocol~\citep{fan2026signalhidenoisespectral}. The main ImageNet and text-to-image results use five independently trained seeds. Ablations use three seeds, except that we reuse the five-seed results for W256/K2 and W256/K4. For ImageNet, FID and Inception Score~\citep{3295222.3295408,3157096.3157346} are computed at 50K images per seed, and compared checkpoints share class ordering and noise draws.

\paragraph{Compute accounting.}
We measure forward cost on one H800 in BF16 with whole-network TorchInductor, counting a multiply--add as two FLOPs (full profiler contract in Table~\ref{tab:formal-contract}). Throughput is the best Heun-50 value over a shared batch sweep.

\paragraph{Compute-matched JiT control.}
JiT-CM/124G keeps JiT-B/8's raw $8\times8$ tokenizer and routes features through a 128-dimensional input bottleneck and 32 context tokens inserted at block 4 of the 12-block trunk; width 560 with 10 heads measures 124.24 GFLOPs under the profiler above. It is an implementation-defined control matched on forward GFLOPs rather than a published JiT size. The ImageNet-512 control JiT-CM/141G uses the corresponding $16\times16$ tokenizer at width 600 (141.21 GFLOPs); we write JiT-CM for the resolution-matched control in what follows.

\subsection{ImageNet Results}
\label{sec:main-results}
\label{sec:imagenet512}

\begin{table}[t]
\caption{ImageNet quality and compute. SpecSlot rows report the best swept $B$ and CFG per allocation (SF-B/32 at ImageNet-512: CFG 9.50). GFLOPs are conditional forward costs; img/s is peak Heun-50 throughput; -- indicates that SpecSlot's $B$ parameter does not apply. SpecSlot gives two nondominated operating points between JiT's low- and high-cost settings, and its margin over the compute-matched control is larger at ImageNet-512 than at ImageNet-256 in these comparisons. JiT models are retrained for 120 epochs under the aligned recipe.}
\label{tab:main-imagenet}
\centering
\footnotesize
\setlength{\tabcolsep}{5pt}
\begin{tabular}{lrrrrrrr}
\toprule
Model & Tokens & $B$ & CFG & FID $\downarrow$ & IS $\uparrow$ &
GFLOPs $\downarrow$ & img/s $\uparrow$ \\
\midrule
\multicolumn{8}{l}{\textit{ImageNet-256}} \\
JiT-B/16 & 256 & -- & 2.90 & 7.51 & 174.81 & 50.45 & 33.08 \\
JiT-B/8  & 1,024 & -- & 2.90 & 4.34 & 225.72 & 218.50 & 8.15 \\
SF-B/16 & 256 & -- & 2.90 & 7.81 & 172.35 & 50.85 & 32.89 \\
JiT-CM & 1,024 & -- & 2.90 & 6.68 & 182.56 & 124.24 & 8.94 \\
\midrule
\textbf{SpecSlot (W256/K2)} & 512 & 512 & 5.00 & 7.27 & 219.49 & 82.63 & 21.32 \\
\textbf{SpecSlot (W256/K4)} & 1,024 & 1,536 & 4.00 & 6.59 & 215.20 & 129.20 & 9.06 \\
\midrule
\multicolumn{8}{l}{\textit{ImageNet-512}} \\
JiT-B/32 & 256 & -- & 2.90 & 15.88 & 162.12 & 51.50 & 30.61 \\
JiT-B/16 & 1,024 & -- & 2.90 & 8.45 & 196.61 & 219.56 & 7.93 \\
SF-B/32 & 256 & -- & 9.50 & 11.41 & 172.48 & 54.72 & 30.88 \\
JiT-CM & 1,024 & -- & 2.90 & 11.98 & 178.72 & 141.21 & 8.51 \\
\midrule
\textbf{SpecSlot (W256/K2)} & 512 & 3,072 & 10.50 & 10.94 & 183.69 & 94.52 & 18.32 \\
\textbf{SpecSlot (W256/K4)} & 1,024 & 4,096 & 8.25 & 9.14 & 208.24 & 135.64 & 12.69 \\
\bottomrule
\end{tabular}
\end{table}

On ImageNet-256, SpecSlot-K2 and SpecSlot-K4 are two measured, nondominated operating points between JiT-B/16 and JiT-B/8 (Table~\ref{tab:main-imagenet}): K2 already improves over JiT-B/16, and K4 improves further while using 40.9\% fewer FLOPs than JiT-B/8, which remains the quality leader. JiT-CM gives the closer compute comparison: SpecSlot-K4 matches its mean FID within the five-seed 95\% interval at nearly identical FLOPs and throughput (Appendix~\ref{app:five-seed}), and descends faster at early-to-mid checkpoints (Figure~\ref{fig:training-convergence}).

At ImageNet-512, each anchor covers four times as many pixels as at ImageNet-256, while our global sequence stays at 288 tokens. Larger spectral windows and the spatial correction can accommodate the additional within-anchor detail without adding global communication, offering a possible explanation for SpecSlot's larger margin at this resolution. SpecSlot-K4 improves FID by 2.84 over JiT-CM at slightly lower cost, and is only 0.69 FID behind JiT-B/16 at 38.2\% lower cost and $1.6\times$ higher throughput; it also attains the highest Inception Score in the panel, and both allocations outperform the input-side spectral baseline SF-B/32. The lower-cost SpecSlot-K2 stays within 1.04 FID of the control at 33.1\% less compute.

Against the strongest JiT model in each panel, which uses about $1.7\times$ our compute, our FID deficit narrows from 2.25 at ImageNet-256 to 0.69 at ImageNet-512, while our lead over JiT-CM grows from 0.09 to 2.84. These two operating points are consistent with the within-anchor allocation mechanism above, but do not by themselves isolate the cause of the larger margin.

Paired ImageNet-256 samples are shown in Figure~\ref{fig:qualitative-main}; additional examples are provided in Appendix Figure~\ref{fig:qualitative-comparison}.

\begin{figure}[!tb]
    \centering
    \includegraphics[width=\textwidth]{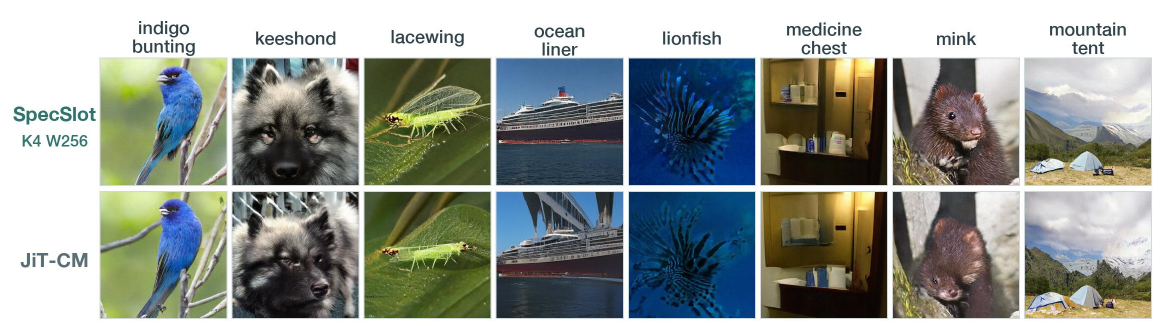}
    \caption{Paired ImageNet-256 generations with the same class label and noise draw: SpecSlot W256/K4 and JiT-CM.}
    \label{fig:qualitative-main}
\end{figure}

\subsection{Text-to-Image Generation}
\label{sec:t2i}

\begin{table}[!t]
\centering
\caption{Text-to-image generation at $256\times256$ after training 400 epochs using the same prompt sets and 50-step Heun sampler. DPG-Bench~\citep{hu2024ellaequipdiffusionmodels} and GenEval~\citep{ghosh2023genevalobjectfocusedframeworkevaluating} report overall scores in percent; forward GFLOPs include the trainable text adapter.}
\label{tab:t2i}
\small
\setlength{\tabcolsep}{0pt}
\begin{tabular*}{0.72\linewidth}{@{}l@{\extracolsep{\fill}}rrrr@{}}
\toprule
Model & CFG & GFLOPs $\downarrow$ & DPG $\uparrow$ & GenEval $\uparrow$ \\
\midrule
JiT-B/16 & 2.9 & 50.71 & 59.22 & 24.89 \\
JiT-B/8 & 2.9 & 218.77 & 61.15 & \textbf{29.78} \\
\midrule
\textbf{SpecSlot (W256/K4)} & 4.0 & 129.86 & \textbf{62.21} & 26.99 \\
\bottomrule
\end{tabular*}
\end{table}

\begin{table}[!t]
\caption{Communication schedules at fixed $W256/K4/B1536$; efficiency columns come from one profiler run.}
\label{tab:routing}
\centering
\small
\setlength{\tabcolsep}{0pt}
\begin{tabular*}{0.72\linewidth}{@{}l@{\extracolsep{\fill}}lcccc@{}}
\toprule
Route & Cross-anchor image states & FID $\downarrow$ & IS $\uparrow$ &
GFLOPs $\downarrow$ & img/s $\uparrow$ \\
\midrule
GGLL & one carrier per anchor & 6.59 & 215.20 & 129.20 & \textbf{9.06} \\
GLGL & one carrier per anchor & 6.95 & 218.95 & 129.20 & 9.00 \\
Dense-global-CM & all image slots & 6.81 & 210.20 & \textbf{128.96} & 5.80 \\
Full-global & all image slots & \textbf{5.84} & \textbf{237.54} & 225.82 & 5.11 \\
\bottomrule
\end{tabular*}
\end{table}

\paragraph{Protocol.}
We test whether the allocation and routing design transfers from class labels to text conditions. We train at $256\times256$ on a mixture of the BLIP3o long-caption and short-caption shards~\citep{chen2025blip3ofamilyfullyopen} and JourneyDB shards~\citep{sun2023journeydbbenchmarkgenerativeimage}. A frozen OpenCLIP ViT-L/14 encoder~\citep{Cherti_2023} and a trainable 32-query adapter provide the denoiser context. SpecSlot reads this context through its carrier states during the global route, while local blocks continue to process image slots; JiT uses the same text encoder and adapter. We retain the ImageNet rectified-flow objective, optimizer, and EMA; the complete training contract is in Appendix Table~\ref{tab:t2i-contract}. Evaluation uses the 50-step Heun sampler with CFG 4.0 for SpecSlot and 2.9 for JiT. Table~\ref{tab:t2i} reports the results.

\paragraph{Results.}
Transferred to text conditioning, SpecSlot attains the highest DPG-Bench score, improving over both JiT baselines at 40.6\% lower forward cost than the stronger JiT-B/8 (Table~\ref{tab:t2i}). On GenEval it improves over JiT-B/16, while JiT-B/8 remains the quality leader at $1.7\times$ the forward cost---the same quality--compute trade-off as on ImageNet-256. This transfer requires no T2I-specific change to the SpecSlot architecture or training objective. Selected text-to-image samples are shown in Figure~\ref{fig:t2i-qualitative-front}; additional examples are provided in Appendix Figure~\ref{fig:t2i-qualitative-appendix}.

%% file: sections/ablation.tex
\begin{figure}[!t]
    \centering
    \includegraphics[width=\textwidth]{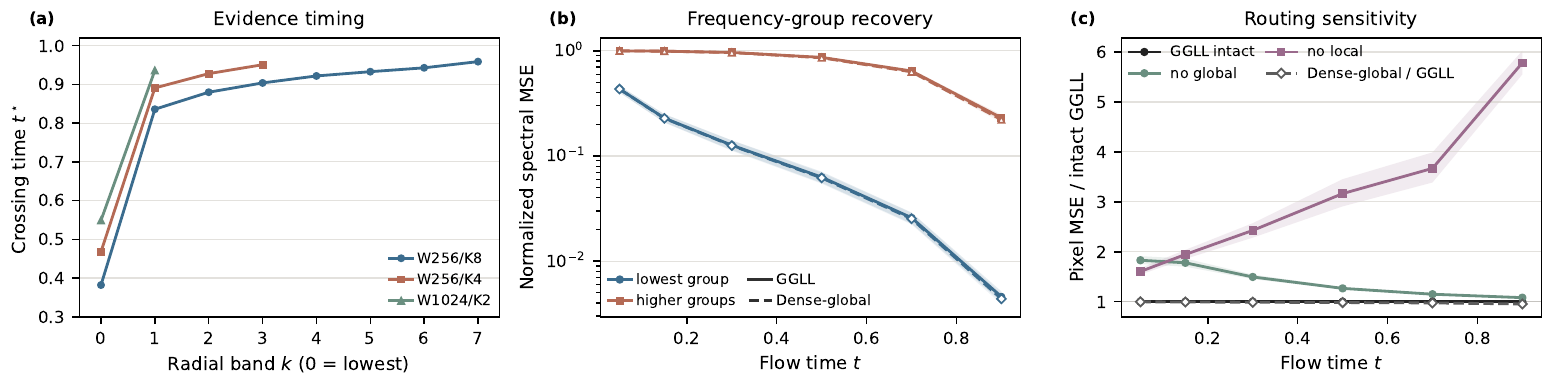}
    \caption{Evidence timing and routed-communication probes. Panels (a)--(c) report the data-to-noise ratio crossing time per radial band for paired W256/K8, W256/K4, and W1024/K2 checkpoints, frequency-group recovery, and route sensitivity. The route interventions remove cross-anchor carrier edges or local cross-band edges inside the shared Transformer while retaining carrier self edges and context reads.}
    \label{fig:mechanism-route}
\end{figure}

\begin{table}[!t]
\centering
\caption{Design analysis at ImageNet-256. (a) Slot allocation: the upper block fixes $N{=}2{,}048$ and the lower fixes $W{=}256$; all rows use $B{=}1536$. (b) Input/output bases at fixed $W256/K4/B1536$.}
\label{tab:allocation}
\begin{minipage}[t]{0.49\linewidth}
\centering
\textbf{(a)}\\[0.35em]
\fontsize{9.0}{10.8}\selectfont
\setlength{\tabcolsep}{0pt}
\renewcommand{\arraystretch}{1.0}
\begin{tabular*}{\linewidth}{@{}l@{\extracolsep{\fill}}cccc@{}}
\toprule
Alloc. & FID $\downarrow$ & IS $\uparrow$ & GFLOPs $\downarrow$ &
img/s $\uparrow$ \\
\midrule
\multicolumn{5}{l}{\textit{Fixed slot budget}} \\
$W64/K32$ & 7.28 & 218.52 & \textbf{195.43} & \textbf{9.00} \\
$W256/K8$ & 6.28 & \textbf{219.52} & 220.02 & 6.55 \\
$W1024/K2$ & \textbf{5.66} & 212.31 & 292.29 & 5.21 \\
\midrule
\multicolumn{5}{l}{\textit{Fixed anchor grid}} \\
$W256/K2$ & 7.39 & \textbf{222.97} & \textbf{83.84} & \textbf{20.10} \\
$W256/K4$ & 6.59 & 215.20 & 129.20 & 9.06 \\
$W256/K8$ & \textbf{6.28} & 219.52 & 220.02 & 6.55 \\
\bottomrule
\end{tabular*}
\end{minipage}\hfill
\begin{minipage}[t]{0.49\linewidth}
\centering
\textbf{(b)}\\[0.35em]
\fontsize{9.0}{10.8}\selectfont
\setlength{\tabcolsep}{0pt}
\renewcommand{\arraystretch}{1.13}
\begin{tabular*}{\linewidth}{@{}l@{\hspace{0.9em}}l@{\extracolsep{\fill}}cccc@{}}
\toprule
Input & Output & FID $\downarrow$ & IS $\uparrow$ & GFLOPs $\downarrow$ &
img/s $\uparrow$ \\
\midrule
\multicolumn{6}{l}{\textit{Single basis}} \\
Freq. & Freq. & 7.34 & \textbf{222.46} & 129.30 & 9.35 \\
Freq. & Spat. & 7.23 & 215.73 & 129.20 & \textbf{9.63} \\
Spat. & Freq. & 7.10 & 221.37 & 129.20 & 9.06 \\
Spat. & Spat. & 6.99 & 215.56 & \textbf{129.10} & 9.34 \\
\midrule
\multicolumn{6}{l}{\textit{Dual basis}} \\
Spat. & Freq.+ Spat. & \textbf{6.59} & 215.20 & 129.20 & 9.06 \\
\bottomrule
\end{tabular*}
\end{minipage}
\end{table}

\Needspace{8\baselineskip}
\section{Ablation Study}
\label{sec:ablation}

\subsection{Understanding the Design}
\label{sec:design-analysis}

\subsubsection{Anchor Density and Coefficient Subdivision}
\label{sec:allocation-results}

Both allocation axes trade compute for quality, but they are not interchangeable: $W$ improves the read, $K$ refines the write (Table~\ref{tab:allocation}). Increasing $K$ at fixed $W$ lowers FID at higher cost with diminishing returns---the last doubling buys only 0.31 FID---while shifting the split toward more anchors at fixed $N$ lowers FID monotonically, with no clear saturation in this range. Within the tested $N{=}2{,}048$ configurations, these trends favor allocating more of the token budget to spatial anchors; we do not claim that this rule generalizes beyond the measured range.

\suppressfloats[t]
\begin{table}[t]
\caption{Effective use of additional spectral slots at fixed $W{=}256$ and evidence width $B{=}1536$. Values are means over paired teacher states. Effective slots/$K$ is the normalized entropy effective rank; definitions are in Appendix~\ref{app:krate-mechanism}.}
\label{tab:mechanism-krate}
\centering
\small
\setlength{\tabcolsep}{0pt}
\begin{tabular*}{\linewidth}{@{}l@{\extracolsep{\fill}}rrrr@{}}
\toprule
$K$ & Velocity MSE & Effective slots/$K$ & Effective local keys & Detail $\to$ carrier \\
\midrule
2 & .10670 & .803 & .729 & .330 \\
4 & .10549 & .620 & .682 & .312 \\
8 & .10497 & .495 & .621 & .275 \\
\bottomrule
\end{tabular*}
\end{table}

At fixed $W$ and $B$, the entropy-effective slot fraction decreases monotonically from 0.803 at $K{=}2$ to 0.495 at $K{=}8$ (Table~\ref{tab:mechanism-krate}). The absolute effective rank still grows, but sublinearly, so token count alone does not measure effective capacity. This utilization pattern is consistent with the diminishing quality gains at larger $K$ and motivates our choice of K2 as the lower-cost point and K4 as the intermediate point.

\subsubsection{Input and Output Representations}
\label{sec:output-ablation}

At the same spectral output, spatial input surpasses frequency input, and the dual-basis hybrid achieves the best FID: 6.59, a further 0.40 below the best single-basis variant at 6.99, without adding tokens or cross-anchor attention edges (Table~\ref{tab:allocation}(b)). This margin indicates that the two output bases are not redundant: the spatial correction supplies structure that the spectral base alone misses. Section~\ref{sec:hybrid-mechanism} localizes this complementarity over the flow and across frequency bands.

\subsubsection{Carrier Routing}
\label{sec:routing-ablation}

To test routing at a fixed forward budget, we train a full-global control (Dense-global-CM) on the same $W256/K4$ configuration, replacing the carrier/local schedule with 12 full-global blocks over all 1,024 image slots; its hidden width is 560, matching the measured GGLL forward cost. GGLL is both better and faster than this control at essentially identical FLOPs (Table~\ref{tab:routing}). Full-global attention reaches a lower FID but at 74.8\% higher cost and lower throughput. The carrier design therefore targets the intermediate, attention-constrained regime, saving 42.8\% of the full-global cost. GLGL keeps the same carrier set and global budget but alternates the global/local order; its FID degrades at identical compute, showing that block order affects performance under this fixed carrier set. Section~\ref{sec:routing-mechanism} provides a possible motivation for early global processing: the lowest-radial role is recovered earliest in the flow.

\subsection{Component Ablations}
\label{sec:component-ablations}

The following controls vary one implementation component at a time at the fixed anchor--slot contract.

\begin{table}[!t]
\centering
\caption{Component ablations at the fixed anchor--slot contract: (a) evidence width; (b) fixed-batch sampling throughput on one H800, in img/s at the indicated Heun interval count (full measurements are in Appendix Table~\ref{tab:sampling-latency}); (c) output-window lapping; (d) input mixing.}
\label{tab:component-ablations}
\begin{minipage}[t]{0.46\linewidth}
\centering
\textbf{(a)}\\[0.35em]
\small
\setlength{\tabcolsep}{0pt}
\begin{tabular*}{\linewidth}{@{}l@{\extracolsep{\fill}}rrrr@{}}
\toprule
Evidence width $B$ & FID $\downarrow$ & IS $\uparrow$ & GFLOPs $\downarrow$ & img/s $\uparrow$ \\
\midrule
$B=1024$ & 6.77 & 221.58 & \textbf{128.19} & \textbf{9.18} \\
$B=1536$ & \textbf{6.59} & 215.20 & 129.20 & 9.06 \\
$B=2048$ & 6.64 & \textbf{224.80} & 129.87 & 9.04 \\
\bottomrule
\end{tabular*}
\end{minipage}\hfill
\begin{minipage}[t]{0.52\linewidth}
\centering
\textbf{(b)}\\[0.35em]
\small
\setlength{\tabcolsep}{0pt}
\begin{tabular*}{\linewidth}{@{}l@{\extracolsep{\fill}}rrr@{}}
\toprule
Route & 20 & 30 & 50 \\
\midrule
GGLL & 21.02 & 13.68 & 8.20 \\
Dense-global-CM & 14.31 & 9.16 & 5.56 \\
Full-global & 12.35 & 8.01 & 4.79 \\
\bottomrule
\end{tabular*}
\end{minipage}

\vspace{10pt}

\begin{minipage}[t]{0.46\linewidth}
\centering
\textbf{(c)}\\[0.35em]
\small
\setlength{\tabcolsep}{0pt}
\begin{tabular*}{\linewidth}{@{}l@{\extracolsep{\fill}}rrrr@{}}
\toprule
Output window & FID $\downarrow$ & IS $\uparrow$ & GFLOPs $\downarrow$ & img/s $\uparrow$ \\
\midrule
Non-overlapping & 6.78 & 211.23 & \textbf{129.19} & \textbf{9.13} \\
50\% overlapping & \textbf{6.59} & \textbf{215.20} & 129.20 & 9.06 \\
\bottomrule
\end{tabular*}
\end{minipage}\hfill
\begin{minipage}[t]{0.52\linewidth}
\centering
\textbf{(d)}\\[0.35em]
\small
\setlength{\tabcolsep}{0pt}
\begin{tabular*}{\linewidth}{@{}l@{\extracolsep{\fill}}rrrr@{}}
\toprule
Input mixer & FID $\downarrow$ & IS $\uparrow$ & GFLOPs $\downarrow$ & img/s $\uparrow$ \\
\midrule
Identity & 6.83 & \textbf{220.38} & \textbf{129.20} & \textbf{9.06} \\
Residual depthwise mixer & \textbf{6.59} & 215.20 & \textbf{129.20} & \textbf{9.06} \\
\bottomrule
\end{tabular*}
\end{minipage}
\end{table}

The best FID among the tested evidence widths is obtained at $B{=}1536$ (Table~\ref{tab:component-ablations}(a)). The residual mixer (Table~\ref{tab:component-ablations}(d)) and 50\% overlap (Table~\ref{tab:component-ablations}(c)) improve FID at nearly unchanged forward cost, supporting their use in the final contract.

\subsection{Sampling Efficiency}
\label{sec:sampling-efficiency}

The main tables report peak Heun-50 throughput. To check how route cost scales with the sampling budget, we also measured fixed-batch BF16 sampling at batch 512 at 20, 30, and 50 Heun intervals (Table~\ref{tab:component-ablations}(b)). GGLL is faster than the compute-matched dense-global control at every tested budget: its throughput is 1.47--1.49$\times$ higher.

%% file: sections/mechanistic.tex
\FloatBarrier
\Needspace{6\baselineskip}
\section{Mechanistic Analysis}
\label{sec:mechanistic-analysis}

\subsection{Spectral Base and Spatial Correction}
\label{sec:hybrid-mechanism}

\begin{figure}[!t]
    \centering
    \includegraphics[width=\textwidth]{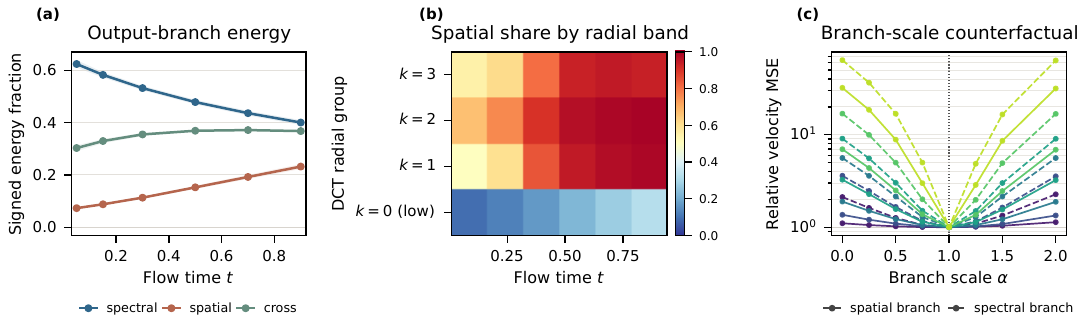}
\caption{Output-branch probes. Panels (a)--(c) show normalized branch self-energies and their signed cross term, spatial self-energy share by radial band, and relative velocity MSE under spatial or spectral scaling over $t\in\{0.05,\ldots,0.90\}$. The energy panel shows 95\% bootstrap intervals; the other panels show means.}
    \label{fig:mechanism-hybrid}
\end{figure}

Figure~\ref{fig:mechanism-hybrid} shows the spectral term has the larger self-energy, while the spatial term's share grows from 0.07 to 0.23 over the flow, and the normalized cross term stays positive---the two terms reinforce, not cancel. Projected through the matched lapped-DCT analysis frame (which does not confer radial coefficient ownership), the spatial term's share is smallest in the lowest radial group and dominates the higher groups late in the flow. Removing either term with the trunk fixed raises velocity MSE: at $t{=}0.90$ by $32.1\times$ (spatial) or $64.1\times$ (spectral), and at $t{=}0.05$ by $1.1\times$ and $2.1\times$. The spectral branch is the main scaffold throughout these probes; the spatial branch supplies an additional late, higher-frequency correction under this parameterization.

\subsection{Carrier Identity and Routed Communication}
\label{sec:routing-mechanism}

\begin{wraptable}{r}{0.42\textwidth}
\vspace{-16pt}
\captionsetup{justification=raggedright,singlelinecheck=false}
\caption{Carrier-identity ablation at fixed $W256/K4/B1536$: only the slot with cross-anchor access changes.}
\label{tab:carrier-assignment}
\centering
\small
\setlength{\tabcolsep}{3pt}
\begin{tabular}{lcc}
\toprule
Global carrier & FID $\downarrow$ & IS $\uparrow$ \\
\midrule
Slot 0 (lowest radial) & \textbf{6.59} & 215.20 \\
Slot 1 & 6.91 & \textbf{220.38} \\
Slot 2 & 7.21 & 203.90 \\
Slot 3 & 6.97 & 214.87 \\
\bottomrule
\end{tabular}
\vspace{-4pt}
\end{wraptable}

We vary only which slot receives cross-anchor access, holding the $W256/K4$ allocation, GGLL graph, and compute fixed (Table~\ref{tab:carrier-assignment}). Replacing the lowest-radial carrier with slot 1, 2, or 3 raises mean FID in every case (Table~\ref{tab:carrier-assignment}): the assignment matters, and the lowest-radial role is also the one whose signal emerges earliest. In Figure~\ref{fig:mechanism-route}, this band is the first to cross $\mathrm{DNR}=1$, motivating the assignment of slot~0 to the global path. Table~\ref{tab:routing} compares separately trained communication graphs, whereas the GGLL intervention curves in Figure~\ref{fig:mechanism-route} change only route edges inside the shared Transformer; Dense-global is a separately trained reference. On paired states, the lowest radial group is recovered earlier than the higher groups; removing cross-anchor carrier edges costs most early, while removing local cross-band edges costs most late.

%% file: sections/conclusion.tex
\section{Conclusion}
\label{sec:conclusion}

SpecSlot recasts tokenization as allocation: a fixed token budget is split between anchor density and spectral subdivision, while global attention is assigned by role. On ImageNet-256, it reaches a comparable final FID to the FLOPs-matched JiT control and descends faster early in training; on ImageNet-512, it improves FID by 2.84 over that control and approaches JiT-B/16 at 38\% lower cost. On text-to-image generation, the same design surpasses JiT-B/8 on DPG-Bench (62.21 vs. 61.15) at 40.6\% lower forward cost. These results show that, in the evaluated pixel-space settings, output detail and global-communication cost can be decoupled through tokenization. Our diagnostics find the lowest-radial role is recovered earliest, supporting its use as the carrier. Allocation gives pixel-space models a new way to trade detail for communication.

%% file: sections/appendix.tex
\section{Implementation Contract}
\label{app:impl}

Table~\ref{tab:formal-contract} records the implementation and training contract of the final class-conditional models, and Table~\ref{tab:t2i-contract} that of the text-to-image model.

\begin{table}[!t]
\caption{Implementation and training contract of the class-conditional $W256/K4$ models. Resource rows are measured on one H800 with BF16 whole-network TorchInductor at profiler batch 128; the ImageNet-512 GFLOPs and throughput are the main-table measurements, and its peak-memory row is the logged training peak per GPU at batch 128.}
\label{tab:formal-contract}
\centering
\begin{minipage}[t]{0.49\textwidth}
\centering
\textbf{ImageNet-256}\par\smallskip
\footnotesize
\renewcommand{\arraystretch}{1.05}
\begin{tabularx}{\linewidth}{@{}>{\raggedright\arraybackslash}p{0.43\linewidth}>{\raggedright\arraybackslash}X@{}}
\toprule
Quantity & Value \\
\midrule
Image side $H$ & 256 \\
Input cell / stride $s$ & 16 / 16 \\
Spatial anchors $W$ & 256 \\
Slots per anchor $K$ & 4 \\
Image slots $N=WK$ & 1,024 \\
Hidden / FFN width / heads / blocks & 768 / 3,072 / 12 / 12 \\
Block design & pre-RMSNorm ($\epsilon{=}10^{-6}$), RMS-normed q/k, SwiGLU FFN \\
Conditioning & sinusoidal-256 $t$ embedding; AdaLN shift/scale/gates from $t$ + 32 class-context slots \\
Evidence width $B$ & 1,536 \\
Input mixer & zero-initialized residual depthwise $3\times3$ \\
Spectral-base window / stride & 32 / 16 \\
Spatial correction tiles / anchor stride & zero-init, row-major $2\times2$ / 16 (K2: $1\times2$, K8: $2\times4$, K32: $4\times8$) \\
Routing & $(G,G,L,L)\times3$ \\
Input DCT / frequency gate & none / none \\
Prediction / loss & image-shaped prediction / pixel velocity MSE \\
Training epochs / GPUs & 120 / 8 \\
Batch per GPU (effective) & 64 (512) \\
Optimizer / schedule & AdamW / constant after warmup \\
Actual learning rate & $1\times10^{-4}$ \\
EMA decays & 0.9999 / 0.9996 \\
Time sampling / label dropout & logit-normal $(-0.8, 0.8^2)$ / 0.1 \\
Reported training seeds & 12345--12349 \\
Trainable parameters & 138.025728M \\
Profiler device / precision & H800 / BF16 \\
Conditional forward GFLOPs & 129.20 \\
Conditional peak memory (batch 128) & 5.56 GB \\
Heun-50 peak memory (batch 128) & 5.84 GB \\
Heun-50 throughput (batch 128) & 9.06 img/s \\
\bottomrule
\end{tabularx}
\end{minipage}\hfill
\begin{minipage}[t]{0.49\textwidth}
\centering
\textbf{ImageNet-512}\par\smallskip
\footnotesize
\renewcommand{\arraystretch}{1.05}
\begin{tabularx}{\linewidth}{@{}>{\raggedright\arraybackslash}p{0.43\linewidth}>{\raggedright\arraybackslash}X@{}}
\toprule
Quantity & Value \\
\midrule
Image side $H$ & 512 \\
Input cell / stride $s$ & 32 / 32 \\
Spatial anchors $W$ & 256 \\
Slots per anchor $K$ & 4 \\
Image slots $N=WK$ & 1,024 \\
Hidden / FFN width / heads / blocks & 768 / 3,072 / 12 / 12 \\
Block design & pre-RMSNorm ($\epsilon{=}10^{-6}$), RMS-normed q/k, SwiGLU FFN \\
Conditioning & sinusoidal-256 $t$ embedding; AdaLN shift/scale/gates from $t$ + 32 class-context slots \\
Evidence width $B$ & 4,096 \\
Input mixer & zero-initialized residual depthwise $3\times3$ \\
Spectral-base window / stride & 64 / 32 \\
Spatial correction tiles / anchor stride & zero-init, row-major $2\times2$ / 32 \\
Routing & $(G,G,L,L)\times3$ \\
Input DCT / frequency gate & none / none \\
Prediction / loss & image-shaped prediction / pixel velocity MSE \\
Training epochs / GPUs & 120 / 8 \\
Batch per GPU (effective) & 128 (1,024) \\
Optimizer / schedule & AdamW / constant after warmup \\
Actual learning rate & $2\times10^{-4}$ \\
EMA decays & 0.9999 / 0.9996 \\
Time sampling / label dropout & logit-normal $(-0.8, 0.8^2)$ / 0.1 \\
Reported training seeds & $\{0,1,2,3,4\}$ \\
Trainable parameters & 160.859904M \\
Profiler device / precision & H800 / BF16 \\
Conditional forward GFLOPs & 135.64 \\
Training peak memory (batch 128 / GPU) & 17.39 GB / GPU \\
Heun-50 throughput (batch 128) & 12.69 img/s \\
\bottomrule
\end{tabularx}
\end{minipage}
\end{table}

\begin{table}[!tbp]
\caption{Text-to-image implementation and training contract (Section~\ref{sec:t2i}).}
\label{tab:t2i-contract}
\centering
\footnotesize
\renewcommand{\arraystretch}{1.05}
\begin{tabularx}{\linewidth}{@{}>{\raggedright\arraybackslash}p{0.43\linewidth}>{\raggedright\arraybackslash}X@{}}
\toprule
Quantity & Value \\
\midrule
Backbone configuration & W256/K4/B1536 at $256\times256$ (Table~\ref{tab:formal-contract}) \\
Text encoder & frozen OpenCLIP ViT-L/14, 77-token states \\
Text adapter & 32-query single-layer cross-attention \\
Training data & BLIP3o long/short-caption + JourneyDB shards \\
Training epochs / GPUs & 400 / 8 (64 images per GPU) \\
Optimizer / learning rate & AdamW / $2\times10^{-4}$ \\
Caption dropout / EMA & 0.1 / 0.9999 \\
Sampler / CFG & Heun-50 / 4.0 (SpecSlot), 2.9 (JiT) \\
Benchmarks & GenEval (553 prompts $\times$ 4), DPG-Bench (1,065 $\times$ 4) \\
\bottomrule
\end{tabularx}
\end{table}

\section{Classifier-free Guidance Sweep}
\label{app:cfg-sweep}

We use JiT-B/8 on ImageNet-256 as a concrete example of the guidance-scale search. Every point evaluates the same epoch-120 EMA1 checkpoint on 50K images with per-rank batch size 512, the legacy-uniform Heun sampler with 50 steps, generation seed 0, and guidance interval $[0.1,1.0]$; only CFG changes. We first run a coarse grid at CFG $\in\{1.0,2.0,2.9,4.0,5.0,6.0,7.0,8.0\}$ (Table~\ref{tab:cfg-sweep}(a)). We then refine the low-FID region with a local $\pm1$ window and increments of 0.25. Table~\ref{tab:cfg-sweep}(b) lists the fine-grid evaluations retained in the archive; the coarse optimum 2.90 is also the fine optimum. We use CFG 2.9 for JiT in the main comparisons, rather than 3.0, to match the scale used by Spectral Forcing~\citep{fan2026signalhidenoisespectral}.

\begin{table}[!htbp]
\centering
\caption{CFG sweeps for JiT-B/8 on ImageNet-256: (a) coarse grid; (b) fine grid around the coarse-grid optimum (the coarse optimum 2.90 included for reference). All points use the same epoch-120 EMA1 checkpoint and evaluation protocol.}
\label{tab:cfg-sweep}
\renewcommand{\arraystretch}{0.90}
\begin{minipage}[t]{0.47\linewidth}
\centering
\textbf{(a)}\\[0.35em]
\small
\setlength{\tabcolsep}{0pt}
\begin{tabular*}{\linewidth}{@{}r@{\extracolsep{\fill}}rr@{}}
\toprule
CFG & FID-50K $\downarrow$ & IS $\uparrow$ \\
\midrule
1.00 & 29.4893 & 53.4115 \\
2.00 & 7.3136 & 154.0057 \\
2.90 & \textbf{4.3579} & 225.1920 \\
4.00 & 5.0462 & 281.4794 \\
5.00 & 6.3617 & 312.9500 \\
6.00 & 7.5583 & 334.1862 \\
7.00 & 8.5149 & 349.8076 \\
8.00 & 9.2679 & 359.5481 \\
\bottomrule
\end{tabular*}
\end{minipage}\hfill
\begin{minipage}[t]{0.47\linewidth}
\centering
\textbf{(b)}\\[0.35em]
\small
\setlength{\tabcolsep}{0pt}
\begin{tabular*}{\linewidth}{@{}r@{\extracolsep{\fill}}rr@{}}
\toprule
CFG & FID-50K $\downarrow$ & IS $\uparrow$ \\
\midrule
2.25 & 5.7930 & 176.5661 \\
2.50 & 4.9153 & 197.2529 \\
2.75 & 4.4765 & 215.3064 \\
2.90 & \textbf{4.3579} & 225.1920 \\
3.25 & 4.3611 & 246.4003 \\
3.50 & 4.5283 & 259.3960 \\
3.75 & 4.7617 & 270.9640 \\
\bottomrule
\end{tabular*}
\end{minipage}
\end{table}

\section{Sampling-step Latency Scaling}
\label{app:sampling-latency}
\enlargethispage{1\baselineskip}

The main tables use 50 Heun intervals for quality evaluation. We additionally measured sampling latency at 20, 30, and 50 intervals for the trained $W256/K4$ model and two routing controls. The measurement uses one H800, BF16, a fixed sampling batch of 512, one warm-up repeat, and three timed repeats. The sampler is the same legacy-uniform Heun implementation used for the quality results. Under the profiler's classifier-free-guidance accounting, the corresponding network-evaluation counts are 73, 113, and 187 per image. We report both quantities to avoid confusing an integration budget with a network-evaluation count. The route-specific CFG values are 4.0 for GGLL, 4.25 for Dense-global-CM, and 3.5 for Full-global.

Table~\ref{tab:sampling-latency} reports the median model-sampling latency. The compute-matched dense-global control has nearly the same per-evaluation FLOPs as GGLL, so it is the primary routing comparison; the full-global row uses a wider network and is included as a cost reference. At every tested budget, GGLL is faster than both controls. Relative to Dense-global-CM, its throughput is 1.47--1.49$\times$ higher across the three budgets; relative to full-global, it is 1.70--1.71$\times$ higher. Figure~\ref{fig:sampling-latency} shows the same scaling in latency.

These measurements isolate fixed-batch model-sampling latency under per-network compilation; the quality numbers in the main tables remain the 50-step evaluations defined in Section~\ref{sec:experimental-setup}.

\begin{table}[!bt]
\caption{Fixed-batch sampling latency on one H800. ``Evals/image'' is the actual network-call count under classifier-free guidance; ``TFLOPs/image'' is the analytic cumulative model-forward compute.}
\label{tab:sampling-latency}
\centering
\footnotesize
\renewcommand{\arraystretch}{0.90}
\setlength{\tabcolsep}{4pt}
\begin{tabular}{llrrrrr}
\toprule
Route & Heun & Evals/image & GFLOPs/eval & TFLOPs/image & ms/image & img/s \\
\midrule
GGLL & 20 & 73  & 128.59 & 9.39  & 47.56  & 21.02 \\
Dense-global-CM & 20 & 73  & 128.96 & 9.41  & 69.89  & 14.31 \\
Full-global & 20 & 73  & 225.22 & 16.44 & 80.99  & 12.35 \\
\midrule
GGLL & 30 & 113 & 128.59 & 14.53 & 73.09  & 13.68 \\
Dense-global-CM & 30 & 113 & 128.96 & 14.57 & 109.20 & 9.16 \\
Full-global & 30 & 113 & 225.22 & 25.45 & 124.83 & 8.01 \\
\midrule
GGLL & 50 & 187 & 128.59 & 24.05 & 121.88 & 8.20 \\
Dense-global-CM & 50 & 187 & 128.96 & 24.11 & 179.95 & 5.56 \\
Full-global & 50 & 187 & 225.22 & 42.12 & 208.60 & 4.79 \\
\bottomrule
\end{tabular}
\end{table}

\begingroup
\setlength{\intextsep}{0pt}
\begin{figure}[H]
    \centering
    \includegraphics[width=0.72\linewidth]{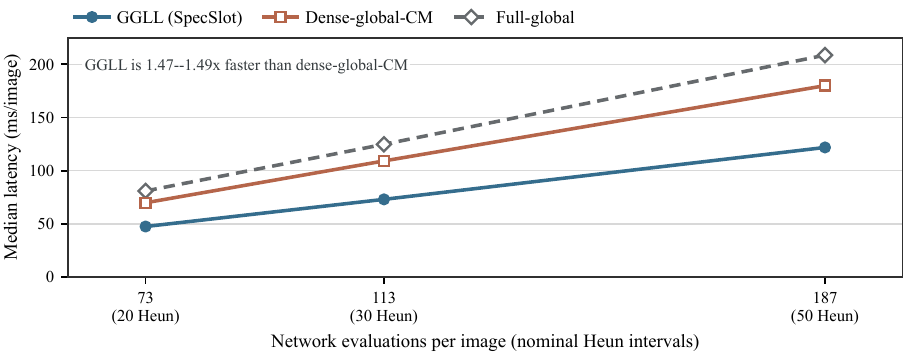}
    \caption{Sampling latency as the evaluation budget increases. The x-axis gives the profiler-observed network evaluations per image; the nominal Heun intervals appear in parentheses.}
    \label{fig:sampling-latency}
\end{figure}
\endgroup
\FloatBarrier

\section{Five-seed ImageNet-256 Results}
\label{app:five-seed}

The main tables report means over independently trained seeds. This section exposes the individual runs and the across-seed uncertainty for the two $W256$ allocations used in the ImageNet-256 comparison. Table~\ref{tab:five-seed-per-seed} lists each training seed, while Table~\ref{tab:five-seed-summary} reports the FID and Inception Score statistics. The training seed identifies the independently trained checkpoint; the generation seed is fixed to zero for every row. Evaluations use epoch-120 EMA1 weights, 50K generated images, the legacy-uniform 50-step Heun sampler, and the same class ordering and noise draws. The selected CFG is 5.0 for K2 and 4.0 for K4, as in Table~\ref{tab:main-imagenet}; the K2 runs use evidence width $B=512$.

\begin{table}[!tbp]
\centering
\begin{minipage}[t]{0.66\linewidth}
\centering
\caption{Per-training-seed ImageNet-256 evaluation on 50K generated images per checkpoint. FID and IS are reported for training seeds 12345--12349.}
\label{tab:five-seed-per-seed}
\small
\setlength{\tabcolsep}{9pt}
\begin{tabular}{llrr}
\toprule
Allocation & Training seed & FID-50K $\downarrow$ & IS $\uparrow$ \\
\midrule
\multirow{5}{*}{$W256/K2$}
 & 12345 & 7.2191 & 214.2811 \\
 & 12346 & 7.0054 & 237.4801 \\
 & 12347 & 7.6025 & 206.7721 \\
 & 12348 & 7.3332 & 226.7448 \\
 & 12349 & 7.1983 & 212.1924 \\
\midrule
\multirow{5}{*}{$W256/K4$}
 & 12345 & 6.4266 & 232.6772 \\
 & 12346 & 6.6197 & 212.0868 \\
 & 12347 & 6.4503 & 224.0468 \\
 & 12348 & 6.8389 & 201.4354 \\
 & 12349 & 6.6250 & 205.7543 \\
\bottomrule
\end{tabular}
\end{minipage}
\end{table}

\begin{table}[!tbp]
\centering
\caption{Across-training-seed statistics for the ImageNet-256 runs of Table~\ref{tab:five-seed-per-seed}: (a) FID; (b) Inception Score. SD is the sample standard deviation, SE is $\mathrm{SD}/\sqrt{n}$, and the interval is the two-sided 95\% Student-$t$ interval with $n-1$ degrees of freedom.}
\label{tab:five-seed-summary}
\begin{minipage}[t]{0.48\linewidth}
\centering
\textbf{(a)}\\[0.35em]
\small
\setlength{\tabcolsep}{0pt}
\begin{tabular*}{\linewidth}{@{}l@{\extracolsep{\fill}}rr@{}}
\toprule
Statistic & $W256/K2$ & $W256/K4$ \\
\midrule
$n$ & 5 & 5 \\
Mean FID & 7.2717 & 6.5921 \\
SD & 0.2192 & 0.1660 \\
SE & 0.0981 & 0.0742 \\
95\% CI & [6.9995,\ 7.5439] & [6.3860,\ 6.7982] \\
Median & 7.2191 & 6.6197 \\
Min--Max & [7.0054,\ 7.6025] & [6.4266,\ 6.8389] \\
\bottomrule
\end{tabular*}
\end{minipage}\hfill
\begin{minipage}[t]{0.48\linewidth}
\centering
\textbf{(b)}\\[0.35em]
\small
\setlength{\tabcolsep}{0pt}
\begin{tabular*}{\linewidth}{@{}l@{\extracolsep{\fill}}rr@{}}
\toprule
Statistic & $W256/K2$ & $W256/K4$ \\
\midrule
$n$ & 5 & 5 \\
Mean IS & 219.4941 & 215.2001 \\
SD & 12.4339 & 12.9624 \\
SE & 5.5606 & 5.7970 \\
95\% CI & [204.0554,\ 234.9328] & [199.1051,\ 231.2951] \\
Median & 214.2811 & 212.0868 \\
Min--Max & [206.7721,\ 237.4801] & [201.4354,\ 232.6772] \\
\bottomrule
\end{tabular*}
\end{minipage}
\end{table}

The aligned five-seed mean trajectories in Figure~\ref{fig:training-convergence}, presented in the main comparison, complement the endpoint statistics above.

\section{Additional Mechanistic Diagnostics}
\label{app:mechanistic-diagnostics}

This appendix collects measurements that probe how SpecSlot uses its explicit roles. Unless stated otherwise, a probe loads a checkpoint and intervenes on its states without updating any parameter; these results characterize representation and communication rather than claim end-to-end quality gains. For the route probe, the intervention is made inside the shared Transformer after slot formation and before the output head, so its interpretation concerns the spectral-slot/GGLL submodule rather than a particular output basis.

\subsection{Evidence timing and route interventions}
\label{app:dnr-routing}

We first measure an empirical data-to-noise ratio (DNR) for each radial coefficient group in the lapped frame. For group $k$, $\mathrm{DNR}_k(t)=t^2\mathbb E\lVert A_kx\rVert^2/[(1-t)^2\mathbb E\lVert A_k\epsilon\rVert^2]$, where $A_k$ denotes the matched lapped-frame analysis restricted to that group. The crossing time $t^\star$ is the first flow time at which the group reaches DNR $\geq 1$; it is a descriptive measure of when the noisy input contains more signal than noise, not a learned gate. Table~\ref{tab:mechanism-dnr} reports the crossings for the paired W256/K8, W256/K4, and W1024/K2 checkpoints.

\begin{table}[!t]
\centering
\begin{minipage}[t]{0.74\linewidth}
\centering
\caption{Empirical DNR frontiers from the lapped-frame probe. Entries list the crossing times $t^\star$ in increasing radial-group order.}
\label{tab:mechanism-dnr}
\small
\setlength{\tabcolsep}{4pt}
\begin{tabular}{@{}ll@{}}
\toprule
Allocation & $t^\star$ for successive radial groups \\
\midrule
$W256/K8$ & $\{0.382,0.836,0.880,0.904,0.922,0.933,0.943,0.959\}$ \\
$W256/K4$ & $\{0.468,0.891,0.928,0.951\}$ \\
$W1024/K2$ & $\{0.549,0.936\}$ \\
\bottomrule
\end{tabular}
\end{minipage}
\end{table}

The frontiers move when $W$ or $K$ changes: the low group becomes usable earlier, while progressively finer groups become usable later. This is consistent with assigning the early cross-anchor path to the lowest radial role. In the matched teacher-forced route intervention, removing the global carrier path primarily increases the low-group and early error; removing local mixing primarily increases detail-group error later in the trajectory. For example, the mid-time log-MSE penalties for the W256/K8 and W256/K4 probes are approximately $0.14/1.66$ and $0.20/2.14$ (global/local). The on-policy intervention has the same ordering: global removal changes the early coarse field (low-frequency change about $0.98$, $0.99$, and $1.05$ for the three allocations), while local removal continues to change mid/high frequency content in the middle and late segments. These measurements are consistent with the role assignment used by GGLL.

\subsection{Allocation and effective use of additional slots}
\label{app:krate-mechanism}

To separate single-step fitting from the use of extra role slots, a paired fixed-$W=256$, fixed-$B=1536$ analysis compares $K\in\{2,4,8\}$. The models share the input, lapped output, and GGLL interfaces used by the registered K-rate diagnostic. For reproducibility, the effective-slot value is the entropy effective rank of each anchor's slot Gram matrix, divided by $K$; the effective-local-key value is $\exp(H_q)/K$ averaged over local queries, with $H_q=-\sum_j p_{qj}\log p_{qj}$; and detail-to-carrier is the mean local attention probability from detail queries to the carrier key. Table~\ref{tab:mechanism-krate} in Section~\ref{sec:allocation-results} shows that although the teacher-forced velocity error changes little, the fraction of effective slots and the fraction of local keys decline as the within-anchor sequence grows.

Thus adding slots can slightly improve teacher-forced fitting without providing the same number of independent, strongly coupled directions. This is consistent with an allocation sweet spot: $K$ controls how finely the coefficient ownership is divided, but it is not a promise that every added slot contributes an equal amount of information.

\section{Paired Qualitative Comparisons}
\label{app:qualitative-comparison}

\subsection{ImageNet examples}
\label{app:imagenet-qualitative}

\begin{figure}[!b]
    \centering
    \includegraphics[width=0.97\textwidth,trim=0 28bp 0 2bp,clip]{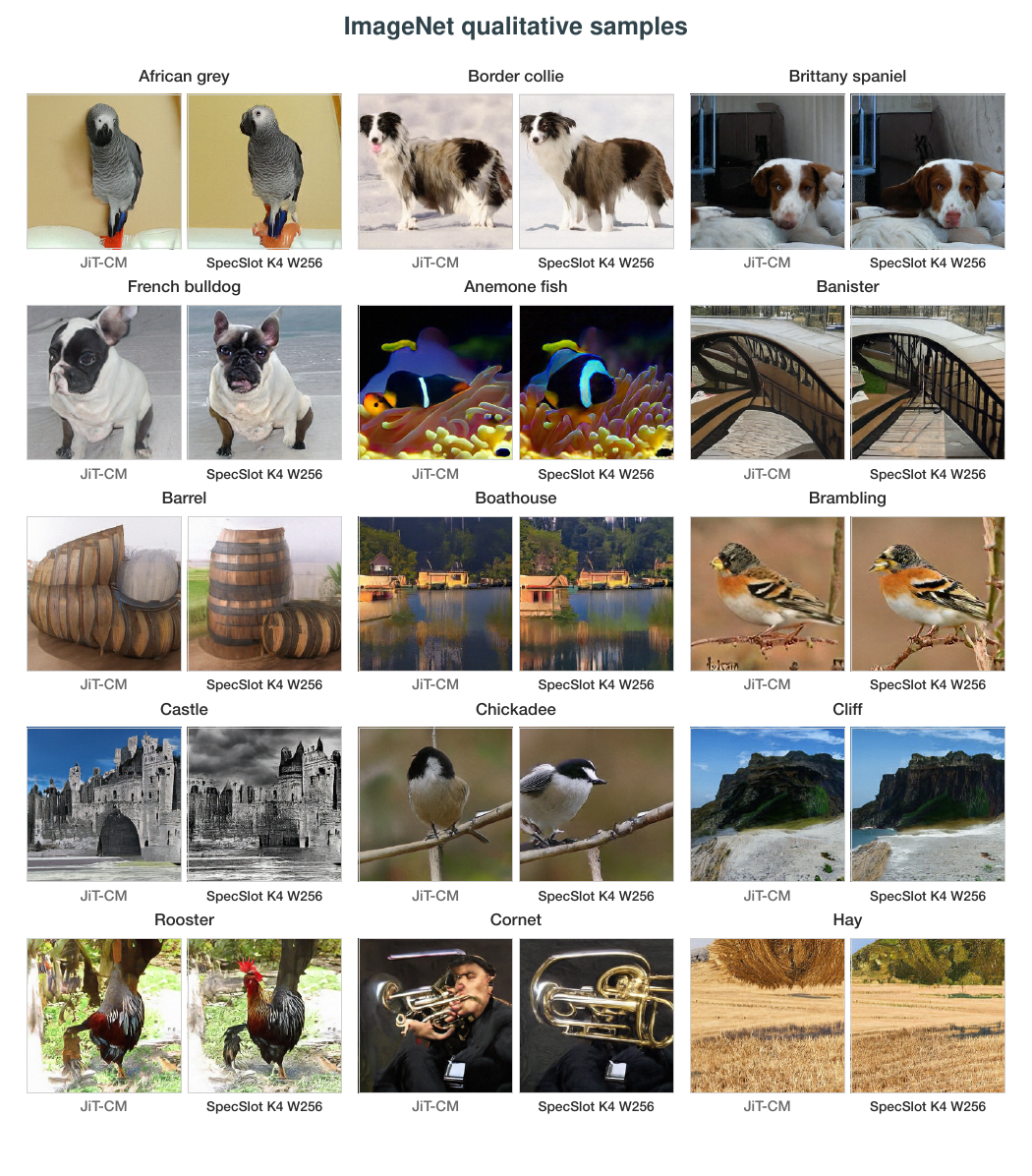}
    \caption{Selected paired generations on ImageNet-256. Within each pair, JiT-CM is shown on the left and SpecSlot W256/K4 on the right; both use the same class and noise draw.}
    \label{fig:qualitative-comparison}
\end{figure}

Figure~\ref{fig:qualitative-comparison} compares selected ImageNet-256 generations from JiT-CM and SpecSlot W256/K4 under the same class and noise draw. In these pairs, SpecSlot renders class-specific fine structure more coherently: feather boundaries are clearer for brambling, chickadee, and rooster; the cornet has more regular tubing and bell geometry; and the boathouse preserves sharper roofline and reflection structure. It also reduces visible local distortions in the anemone fish, banister, and barrel examples, where contours or repeated parts are less coherent in the JiT-CM samples. The hay example shows a more regular field-scale texture.

\subsection{Text-to-image examples}
\label{app:t2i-qualitative}

Figure~\ref{fig:t2i-qualitative-appendix} shows additional SpecSlot samples generated from text prompts in DPG-Bench. The examples cover portraits, creatures, environments, architecture, and interiors; the prompt paired with each sample is included in the figure.

\begin{figure}[p]
    \centering
    \includegraphics[width=\textwidth]{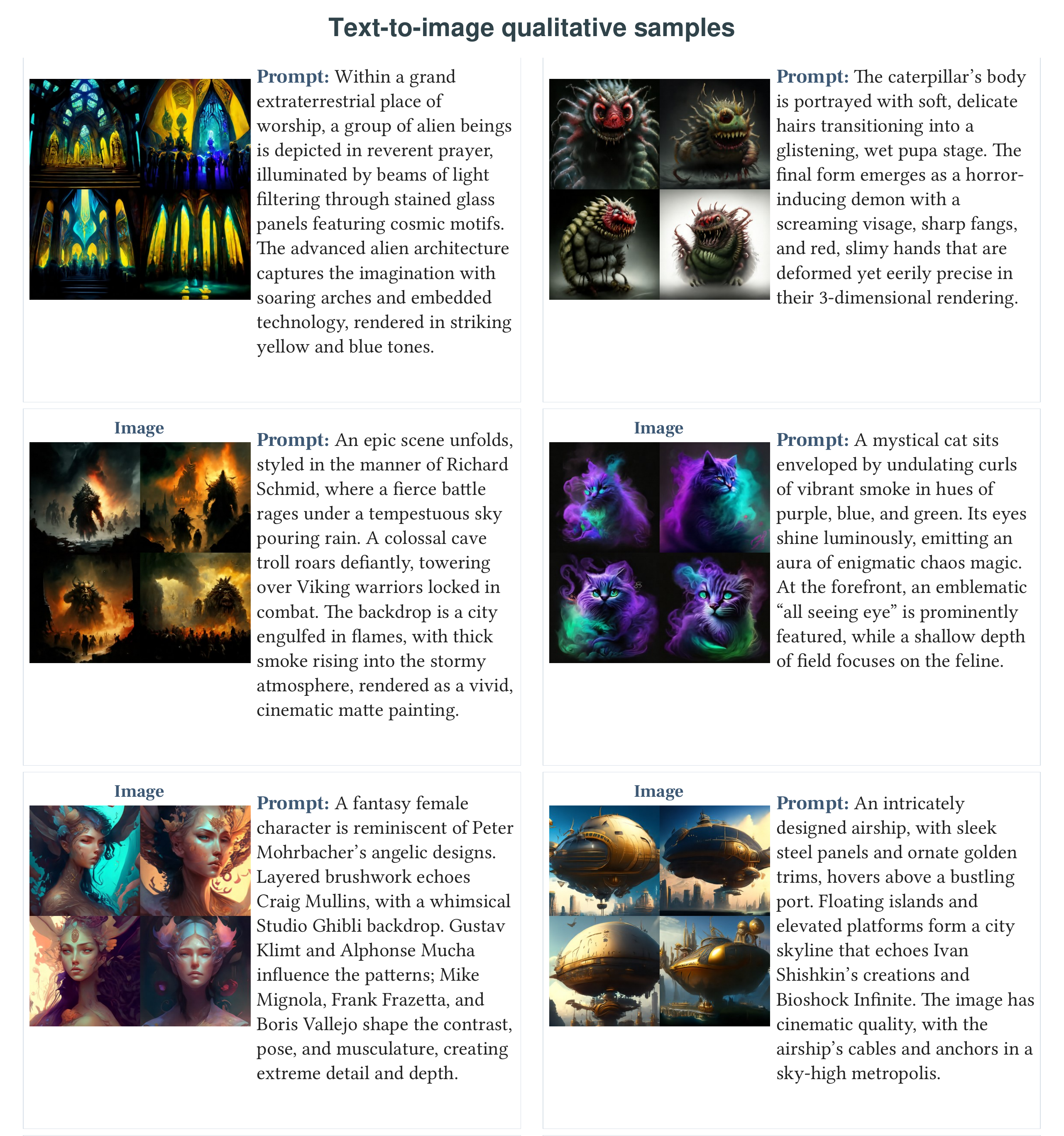}
    \caption{Additional text-to-image samples generated by SpecSlot W256/K4 from DPG-Bench prompts (first part).}
    \label{fig:t2i-qualitative-appendix}
\end{figure}

\begin{figure}[p]
    \ContinuedFloat
    \centering
    \includegraphics[width=\textwidth]{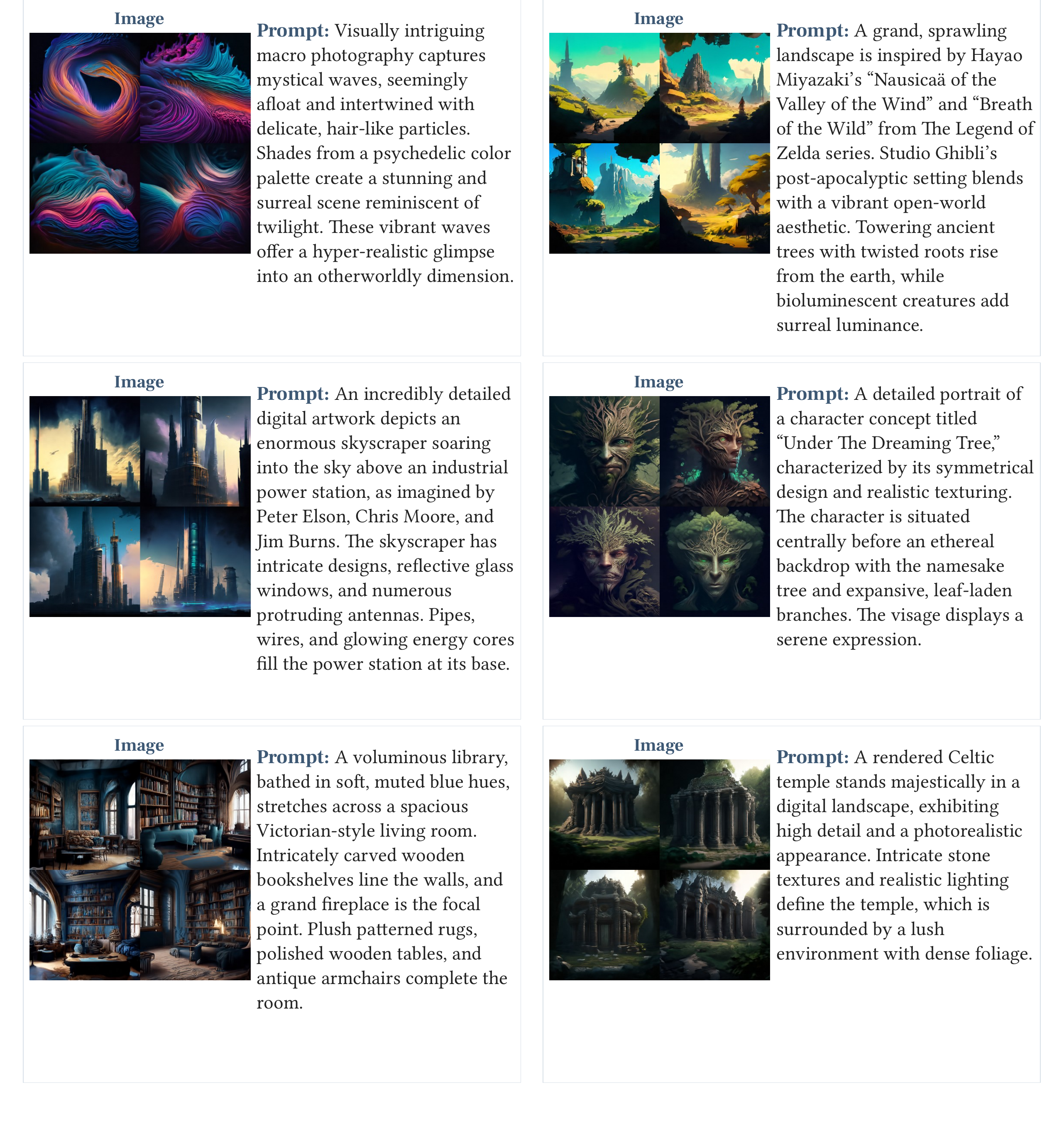}
    \caption[]{Additional text-to-image samples generated by SpecSlot W256/K4 from DPG-Bench prompts (continued).}
\end{figure}

\FloatBarrier

%% file: references.bib
@inproceedings{NEURIPS2021_49ad23d1,
  author    = {Dhariwal, Prafulla and Nichol, Alexander},
  booktitle = {Advances in Neural Information Processing Systems},
  editor    = {M. Ranzato and A. Beygelzimer and Y. Dauphin and P.S. Liang and J. Wortman Vaughan},
  pages     = {8780--8794},
  publisher = {Curran Associates, Inc.},
  title     = {Diffusion Models Beat GANs on Image Synthesis},
  url       = {https://proceedings.neurips.cc/paper_files/paper/2021/file/49ad23d1ec9fa4bd8d77d02681df5cfa-Paper.pdf},
  volume    = {34},
  year      = {2021}
}

@inproceedings{9878449,
  author    = {Rombach, Robin and Blattmann, Andreas and Lorenz, Dominik and Esser, Patrick and Ommer, Björn},
  booktitle = {2022 IEEE/CVF Conference on Computer Vision and Pattern Recognition (CVPR)},
  title     = {High-Resolution Image Synthesis with Latent Diffusion Models},
  year      = {2022},
  volume    = {},
  number    = {},
  pages     = {10674-10685},
  doi       = {10.1109/CVPR52688.2022.01042}
}

@inproceedings{9578911,
  author    = { Esser, Patrick and Rombach, Robin and Ommer, Bjorn },
  booktitle = { 2021 IEEE/CVF Conference on Computer Vision and Pattern Recognition (CVPR) },
  title     = {{ Taming Transformers for High-Resolution Image Synthesis }},
  year      = {2021},
  volume    = {},
  issn      = {},
  pages     = {12868-12878},
  doi       = {10.1109/CVPR46437.2021.01268},
  url       = {https://doi.ieeecomputersociety.org/10.1109/CVPR46437.2021.01268},
  publisher = {IEEE Computer Society},
  address   = {Los Alamitos, CA, USA},
  month     = Jun
}

@inproceedings{10377858,
  author    = { Peebles, William and Xie, Saining },
  booktitle = { 2023 IEEE/CVF International Conference on Computer Vision (ICCV) },
  title     = {{ Scalable Diffusion Models with Transformers }},
  year      = {2023},
  volume    = {},
  issn      = {},
  pages     = {4172-4182},
  doi       = {10.1109/ICCV51070.2023.00387},
  url       = {https://doi.ieeecomputersociety.org/10.1109/ICCV51070.2023.00387},
  publisher = {IEEE Computer Society},
  address   = {Los Alamitos, CA, USA},
  month     = Oct
}

@misc{ma2024sitexploringflowdiffusionbased,
  title         = {SiT: Exploring Flow and Diffusion-based Generative Models with Scalable Interpolant Transformers},
  author        = {Nanye Ma and Mark Goldstein and Michael S. Albergo and Nicholas M. Boffi and Eric Vanden-Eijnden and Saining Xie},
  year          = {2024},
  eprint        = {2401.08740},
  archiveprefix = {arXiv},
  primaryclass  = {cs.CV},
  url           = {https://arxiv.org/abs/2401.08740}
}

@inproceedings{dosovitskiy2021an,
  title     = {An Image is Worth 16x16 Words: Transformers for Image Recognition at Scale},
  author    = {Alexey Dosovitskiy and Lucas Beyer and Alexander Kolesnikov and Dirk Weissenborn and Xiaohua Zhai and Thomas Unterthiner and Mostafa Dehghani and Matthias Minderer and Georg Heigold and Sylvain Gelly and Jakob Uszkoreit and Neil Houlsby},
  booktitle = {International Conference on Learning Representations},
  year      = {2021},
  url       = {https://openreview.net/forum?id=YicbFdNTTy}
}

@misc{hoogeboom2023simplediffusionendtoenddiffusion,
  title         = {Simple diffusion: End-to-end diffusion for high resolution images},
  author        = {Emiel Hoogeboom and Jonathan Heek and Tim Salimans},
  year          = {2023},
  eprint        = {2301.11093},
  archiveprefix = {arXiv},
  primaryclass  = {cs.CV},
  url           = {https://arxiv.org/abs/2301.11093}
}

@misc{bao2023worthwordsvitbackbone,
  title         = {All are Worth Words: A ViT Backbone for Diffusion Models},
  author        = {Fan Bao and Shen Nie and Kaiwen Xue and Yue Cao and Chongxuan Li and Hang Su and Jun Zhu},
  year          = {2023},
  eprint        = {2209.12152},
  archiveprefix = {arXiv},
  primaryclass  = {cs.CV},
  url           = {https://arxiv.org/abs/2209.12152}
}

@misc{jabri2023scalableadaptivecomputationiterative,
  title         = {Scalable Adaptive Computation for Iterative Generation},
  author        = {Allan Jabri and David Fleet and Ting Chen},
  year          = {2023},
  eprint        = {2212.11972},
  archiveprefix = {arXiv},
  primaryclass  = {cs.LG},
  url           = {https://arxiv.org/abs/2212.11972}
}

@misc{li2026basicsletdenoisinggenerative,
  title         = {Back to Basics: Let Denoising Generative Models Denoise},
  author        = {Tianhong Li and Kaiming He},
  year          = {2026},
  eprint        = {2511.13720},
  archiveprefix = {arXiv},
  primaryclass  = {cs.CV},
  url           = {https://arxiv.org/abs/2511.13720}
}

@misc{chen2025pixelflowpixelspacegenerativemodels,
  title         = {PixelFlow: Pixel-Space Generative Models with Flow},
  author        = {Shoufa Chen and Chongjian Ge and Shilong Zhang and Peize Sun and Ping Luo},
  year          = {2025},
  eprint        = {2504.07963},
  archiveprefix = {arXiv},
  primaryclass  = {cs.CV},
  url           = {https://arxiv.org/abs/2504.07963}
}

@inproceedings{Yu_2026_CVPR,
  author    = {Yu, Yongsheng and Xiong, Wei and Nie, Weili and Sheng, Yichen and Liu, Shiqiu and Luo, Jiebo},
  title     = {PixelDiT: Pixel Diffusion Transformers for Image Generation},
  booktitle = {Proceedings of the IEEE/CVF Conference on Computer Vision and Pattern Recognition (CVPR)},
  month     = {June},
  year      = {2026},
  pages     = {14273-14282}
}

@misc{zheng2025diffusiontransformersrepresentationautoencoders,
  title         = {Diffusion Transformers with Representation Autoencoders},
  author        = {Boyang Zheng and Nanye Ma and Shengbang Tong and Saining Xie},
  year          = {2025},
  eprint        = {2510.11690},
  archiveprefix = {arXiv},
  primaryclass  = {cs.CV},
  url           = {https://arxiv.org/abs/2510.11690}
}

@inproceedings{rissanen2023generative,
  title     = {Generative Modelling with Inverse Heat Dissipation},
  author    = {Severi Rissanen and Markus Heinonen and Arno Solin},
  booktitle = {The Eleventh International Conference on Learning Representations },
  year      = {2023},
  url       = {https://openreview.net/forum?id=4PJUBT9f2Ol}
}

@inproceedings{NEURIPS2022_03474669,
  author    = {Guth, Florentin and Coste, Simon and De Bortoli, Valentin and Mallat, Stephane},
  booktitle = {Advances in Neural Information Processing Systems},
  doi       = {10.52202/068431-0035},
  editor    = {S. Koyejo and S. Mohamed and A. Agarwal and D. Belgrave and K. Cho and A. Oh},
  pages     = {478--491},
  publisher = {Curran Associates, Inc.},
  title     = {Wavelet Score-Based Generative Modeling},
  url       = {https://proceedings.neurips.cc/paper_files/paper/2022/file/03474669b759f6d38cdca6fb4eb905f4-Paper-Conference.pdf},
  volume    = {35},
  year      = {2022}
}

@inproceedings{10204642,
  author    = {Phung, Hao and Dao, Quan and Tran, Anh},
  booktitle = {2023 IEEE/CVF Conference on Computer Vision and Pattern Recognition (CVPR)},
  title     = {Wavelet Diffusion Models are fast and scalable Image Generators},
  year      = {2023},
  volume    = {},
  number    = {},
  pages     = {10199-10208},
  doi       = {10.1109/CVPR52729.2023.00983}
}

@inproceedings{ning2025dctdiff,
  title     = {{DCT}diff: Intriguing Properties of Image Generative Modeling in the {DCT} Space},
  author    = {Mang Ning and Mingxiao Li and Jianlin Su and Jia Haozhe and Lanmiao Liu and Martin Benes and Wenshuo Chen and Albert Ali Salah and Itir Onal Ertugrul},
  booktitle = {Forty-second International Conference on Machine Learning},
  year      = {2025},
  url       = {https://openreview.net/forum?id=vDoAA8xKXL}
}

@inproceedings{9710849,
  author    = {Jiang, Liming and Dai, Bo and Wu, Wayne and Loy, Chen Change},
  booktitle = {2021 IEEE/CVF International Conference on Computer Vision (ICCV)},
  title     = {Focal Frequency Loss for Image Reconstruction and Synthesis},
  year      = {2021},
  volume    = {},
  number    = {},
  pages     = {13899-13909},
  doi       = {10.1109/ICCV48922.2021.01366}
}

@misc{falck2025fourierspaceperspectivediffusion,
  title         = {A Fourier Space Perspective on Diffusion Models},
  author        = {Fabian Falck and Teodora Pandeva and Kiarash Zahirnia and Rachel Lawrence and Richard Turner and Edward Meeds and Javier Zazo and Sushrut Karmalkar},
  year          = {2025},
  eprint        = {2505.11278},
  archiveprefix = {arXiv},
  primaryclass  = {stat.ML},
  url           = {https://arxiv.org/abs/2505.11278}
}

@misc{fan2026signalhidenoisespectral,
  title         = {Show the Signal, Hide the Noise: Spectral Forcing for Pixel-Space Diffusion},
  author        = {Weichen Fan and Haiwen Diao and Penghao Wu and Ziwei Liu},
  year          = {2026},
  eprint        = {2606.15236},
  archiveprefix = {arXiv},
  primaryclass  = {cs.CV},
  url           = {https://arxiv.org/abs/2606.15236}
}

@inproceedings{baade2026latent,
  title     = {Latent Forcing: Reordering the Diffusion Trajectory for Pixel-Space Image Generation},
  author    = {Alan Baade and Eric Ryan Chan and Kyle Sargent and Changan Chen and Justin Johnson and Ehsan Adeli and Li Fei-Fei},
  booktitle = {Forty-third International Conference on Machine Learning},
  year      = {2026},
  url       = {https://openreview.net/forum?id=LFUaHUCO1a}
}

@inproceedings{du2026frequencyforcing,
  title     = {Frequency-Forcing: From Scaling-as-Time to Soft Frequency Guidance},
  author    = {Weitao Du},
  booktitle = {ICML 2026 Workshop on Structured Probabilistic Inference {\&} Generative Modeling},
  year      = {2026},
  url       = {https://openreview.net/forum?id=dvTYdwKbLj}
}

@inproceedings{dehghani2023patch,
  title     = {Patch n{\textquoteright} Pack: NaViT, a Vision Transformer for any Aspect Ratio and Resolution},
  author    = {Mostafa Dehghani and Basil Mustafa and Josip Djolonga and Jonathan Heek and Matthias Minderer and Mathilde Caron and Andreas Peter Steiner and Joan Puigcerver and Robert Geirhos and Ibrahim Alabdulmohsin and Avital Oliver and Piotr Padlewski and Alexey A. Gritsenko and Mario Lucic and Neil Houlsby},
  booktitle = {Thirty-seventh Conference on Neural Information Processing Systems},
  year      = {2023},
  url       = {https://openreview.net/forum?id=VpGFHmI7e5}
}

@inproceedings{10208676,
  author    = {Bolya, Daniel and Hoffman, Judy},
  booktitle = {2023 IEEE/CVF Conference on Computer Vision and Pattern Recognition Workshops (CVPRW)},
  title     = {Token Merging for Fast Stable Diffusion},
  year      = {2023},
  volume    = {},
  number    = {},
  pages     = {4599-4603},
  doi       = {10.1109/CVPRW59228.2023.00484}
}

@misc{lee2026variablelengthtokenizationlearnableglobal,
  title         = {Variable-Length Tokenization via Learnable Global Merging for Diffusion Transformers},
  author        = {Dong Hoon Lee and Seunghoon Hong},
  year          = {2026},
  eprint        = {2606.20076},
  archiveprefix = {arXiv},
  primaryclass  = {cs.CV},
  url           = {https://arxiv.org/abs/2606.20076}
}

@inproceedings{chang2025sparsedit,
  title     = {SparseDiT: Token Sparsification for Efficient Diffusion Transformer},
  author    = {Shuning Chang and Pichao WANG and Jiasheng Tang and Fan Wang and Yi Yang},
  booktitle = {The Thirty-ninth Annual Conference on Neural Information Processing Systems},
  year      = {2025},
  url       = {https://openreview.net/forum?id=jTBxyQempF}
}

@article{3586589.3586636,
  author     = {Ho, Jonathan and Saharia, Chitwan and Chan, William and Fleet, David J. and Norouzi, Mohammad and Salimans, Tim},
  title      = {Cascaded diffusion models for high fidelity image generation},
  year       = {2022},
  issue_date = {January 2022},
  publisher  = {JMLR.org},
  volume     = {23},
  number     = {1},
  issn       = {1532-4435},
  journal    = {J. Mach. Learn. Res.},
  month      = jan,
  articleno  = {47},
  numpages   = {33}
}

@inproceedings{tian2024visual,
  title     = {Visual Autoregressive Modeling: Scalable Image Generation via Next-Scale Prediction},
  author    = {Keyu Tian and Yi Jiang and Zehuan Yuan and BINGYUE PENG and Liwei Wang},
  booktitle = {The Thirty-eighth Annual Conference on Neural Information Processing Systems},
  year      = {2024},
  url       = {https://openreview.net/forum?id=gojL67CfS8}
}

@misc{liu2022flowstraightfastlearning,
  title         = {Flow Straight and Fast: Learning to Generate and Transfer Data with Rectified Flow},
  author        = {Xingchao Liu and Chengyue Gong and Qiang Liu},
  year          = {2022},
  eprint        = {2209.03003},
  archiveprefix = {arXiv},
  primaryclass  = {cs.LG},
  url           = {https://arxiv.org/abs/2209.03003}
}

@inproceedings{5206848,
  author    = {Deng, Jia and Dong, Wei and Socher, Richard and Li, Li-Jia and Kai Li and Li Fei-Fei},
  booktitle = {2009 IEEE Conference on Computer Vision and Pattern Recognition},
  title     = {ImageNet: A large-scale hierarchical image database},
  year      = {2009},
  volume    = {},
  number    = {},
  pages     = {248-255},
  doi       = {10.1109/CVPR.2009.5206848}
}

@inproceedings{3295222.3295408,
  author    = {Heusel, Martin and Ramsauer, Hubert and Unterthiner, Thomas and Nessler, Bernhard and Hochreiter, Sepp},
  title     = {GANs trained by a two time-scale update rule converge to a local nash equilibrium},
  year      = {2017},
  isbn      = {9781510860964},
  publisher = {Curran Associates Inc.},
  address   = {Red Hook, NY, USA},
  booktitle = {Proceedings of the 31st International Conference on Neural Information Processing Systems},
  pages     = {6629--6640},
  numpages  = {12},
  location  = {Long Beach, California, USA},
  series    = {NIPS'17}
}

@inproceedings{3157096.3157346,
  author    = {Salimans, Tim and Goodfellow, Ian and Zaremba, Wojciech and Cheung, Vicki and Radford, Alec and Chen, Xi},
  title     = {Improved techniques for training GANs},
  year      = {2016},
  isbn      = {9781510838819},
  publisher = {Curran Associates Inc.},
  address   = {Red Hook, NY, USA},
  booktitle = {Proceedings of the 30th International Conference on Neural Information Processing Systems},
  pages     = {2234--2242},
  numpages  = {9},
  location  = {Barcelona, Spain},
  series    = {NIPS'16}
}

@misc{wang2025pixnerdpixelneuralfield,
  title         = {PixNerd: Pixel Neural Field Diffusion},
  author        = {Shuai Wang and Ziteng Gao and Chenhui Zhu and Weilin Huang and Limin Wang},
  year          = {2025},
  eprint        = {2507.23268},
  archiveprefix = {arXiv},
  primaryclass  = {cs.CV},
  url           = {https://arxiv.org/abs/2507.23268}
}

@inproceedings{3692070.3692447,
  author    = {Crowson, Katherine and Baumann, Stefan Andreas and Birch, Alex and Abraham, Tanishq Mathew and Kaplan, Daniel Z. and Shippole, Enrico},
  title     = {Scalable high-resolution pixel-space image synthesis with hourglass diffusion transformers},
  year      = {2024},
  publisher = {JMLR.org},
  booktitle = {Proceedings of the 41st International Conference on Machine Learning},
  articleno = {377},
  numpages  = {26},
  location  = {Vienna, Austria},
  series    = {ICML'24}
}

@misc{chen2026diptamingdiffusionmodels,
  title         = {DiP: Taming Diffusion Models in Pixel Space},
  author        = {Zhennan Chen and Junwei Zhu and Xu Chen and Jiangning Zhang and Xiaobin Hu and Hanzhen Zhao and Chengjie Wang and Jian Yang and Ying Tai},
  year          = {2026},
  eprint        = {2511.18822},
  archiveprefix = {arXiv},
  primaryclass  = {cs.CV},
  url           = {https://arxiv.org/abs/2511.18822}
}

@misc{he2026hyperdithyperconnectedtransformershighfidelity,
  title         = {HyperDiT: Hyper-Connected Transformers for High-Fidelity Pixel-Space Diffusion},
  author        = {Yu He and Lichen Ma and Zipeng Guo and Xinyuan Shan and Jingling Fu and Dong Chen and Junshi Huang and Yan Li},
  year          = {2026},
  eprint        = {2605.15741},
  archiveprefix = {arXiv},
  primaryclass  = {cs.CV},
  url           = {https://arxiv.org/abs/2605.15741}
}

@misc{huang2025nfigmultiscaleautoregressiveimage,
  title         = {NFIG: Multi-Scale Autoregressive Image Generation via Frequency Ordering},
  author        = {Zhihao Huang and Xi Qiu and Yukuo Ma and Yifu Zhou and Junjie Chen and Hongyuan Zhang and Chi Zhang and Xuelong Li},
  year          = {2025},
  eprint        = {2503.07076},
  archiveprefix = {arXiv},
  primaryclass  = {cs.CV},
  url           = {https://arxiv.org/abs/2503.07076}
}

@misc{nash2021generatingimagessparserepresentations,
  title         = {Generating Images with Sparse Representations},
  author        = {Charlie Nash and Jacob Menick and Sander Dieleman and Peter W. Battaglia},
  year          = {2021},
  eprint        = {2103.03841},
  archiveprefix = {arXiv},
  primaryclass  = {cs.CV},
  url           = {https://arxiv.org/abs/2103.03841}
}

@misc{phung2025dimsumdiffusionmamba,
  title         = {DiMSUM: Diffusion Mamba -- A Scalable and Unified Spatial-Frequency Method for Image Generation},
  author        = {Hao Phung and Quan Dao and Trung Dao and Hoang Phan and Dimitris Metaxas and Anh Tran},
  year          = {2025},
  eprint        = {2411.04168},
  archiveprefix = {arXiv},
  primaryclass  = {cs.CV},
  url           = {https://arxiv.org/abs/2411.04168}
}

@misc{zhao2026dyditdiffusiontransformerstimestep,
  title         = {DyDiT++: Diffusion Transformers with Timestep and Spatial Dynamics for Efficient Visual Generation},
  author        = {Wangbo Zhao and Yizeng Han and Jiasheng Tang and Kai Wang and Hao Luo and Yibing Song and Gao Huang and Fan Wang and Yang You},
  year          = {2026},
  eprint        = {2504.06803},
  archiveprefix = {arXiv},
  primaryclass  = {cs.CV},
  url           = {https://arxiv.org/abs/2504.06803}
}

@misc{jaegle2021perceivergeneralperceptioniterative,
  title         = {Perceiver: General Perception with Iterative Attention},
  author        = {Andrew Jaegle and Felix Gimeno and Andrew Brock and Andrew Zisserman and Oriol Vinyals and Joao Carreira},
  year          = {2021},
  eprint        = {2103.03206},
  archiveprefix = {arXiv},
  primaryclass  = {cs.CV},
  url           = {https://arxiv.org/abs/2103.03206}
}

@inproceedings{9710580,
  author    = { Liu, Ze and Lin, Yutong and Cao, Yue and Hu, Han and Wei, Yixuan and Zhang, Zheng and Lin, Stephen and Guo, Baining },
  booktitle = { 2021 IEEE/CVF International Conference on Computer Vision (ICCV) },
  title     = {{ Swin Transformer: Hierarchical Vision Transformer using Shifted Windows }},
  year      = {2021},
  volume    = {},
  issn      = {},
  pages     = {9992-10002},
  doi       = {10.1109/ICCV48922.2021.00986},
  url       = {https://doi.ieeecomputersociety.org/10.1109/ICCV48922.2021.00986},
  publisher = {IEEE Computer Society},
  address   = {Los Alamitos, CA, USA},
  month     = Oct
}

@misc{beltagy2020longformerlongdocumenttransformer,
  title         = {Longformer: The Long-Document Transformer},
  author        = {Iz Beltagy and Matthew E. Peters and Arman Cohan},
  year          = {2020},
  eprint        = {2004.05150},
  archiveprefix = {arXiv},
  primaryclass  = {cs.CL},
  url           = {https://arxiv.org/abs/2004.05150}
}

@misc{zhou2026trainableloglinearsparseattention,
  title         = {Trainable Log-linear Sparse Attention for Efficient Diffusion Transformers},
  author        = {Yifan Zhou and Zeqi Xiao and Tianyi Wei and Shuai Yang and Xingang Pan},
  year          = {2026},
  eprint        = {2512.16615},
  archiveprefix = {arXiv},
  primaryclass  = {cs.CV},
  url           = {https://arxiv.org/abs/2512.16615}
}

@misc{you2026advancedpixeldiffusionmodel,
  title         = {Advanced Pixel Diffusion Model with Guided Sparse Global Refinement},
  author        = {Weiyi You and Jinhua Zhang and Xingyu Zhou and Wei Long and Junyu Lou and Shuhang Gu},
  year          = {2026},
  eprint        = {2609.00798},
  archiveprefix = {arXiv},
  primaryclass  = {cs.CV},
  url           = {https://arxiv.org/abs/2609.00798}
}

@misc{ma2026decofrequencydecoupledpixeldiffusion,
  title         = {DeCo: Frequency-Decoupled Pixel Diffusion for End-to-End Image Generation},
  author        = {Zehong Ma and Longhui Wei and Shuai Wang and Shiliang Zhang and Qi Tian},
  year          = {2026},
  eprint        = {2511.19365},
  archiveprefix = {arXiv},
  primaryclass  = {cs.CV},
  url           = {https://arxiv.org/abs/2511.19365}
}

@misc{anagnostidis2025flexiditdiffusiontransformereasily,
  title         = {FlexiDiT: Your Diffusion Transformer Can Easily Generate High-Quality Samples with Less Compute},
  author        = {Sotiris Anagnostidis and Gregor Bachmann and Yeongmin Kim and Jonas Kohler and Markos Georgopoulos and Artsiom Sanakoyeu and Yuming Du and Albert Pumarola and Ali Thabet and Edgar Schönfeld},
  year          = {2025},
  eprint        = {2502.20126},
  archiveprefix = {arXiv},
  primaryclass  = {cs.LG},
  url           = {https://arxiv.org/abs/2502.20126}
}

@inproceedings{3495724.3496298,
  author    = {Ho, Jonathan and Jain, Ajay and Abbeel, Pieter},
  title     = {Denoising diffusion probabilistic models},
  year      = {2020},
  isbn      = {9781713829546},
  publisher = {Curran Associates Inc.},
  address   = {Red Hook, NY, USA},
  booktitle = {Proceedings of the 34th International Conference on Neural Information Processing Systems},
  articleno = {574},
  numpages  = {12},
  location  = {Vancouver, BC, Canada},
  series    = {NIPS '20}
}

@misc{song2021scorebasedgenerativemodelingstochastic,
  title         = {Score-Based Generative Modeling through Stochastic Differential Equations},
  author        = {Yang Song and Jascha Sohl-Dickstein and Diederik P. Kingma and Abhishek Kumar and Stefano Ermon and Ben Poole},
  year          = {2021},
  eprint        = {2011.13456},
  archiveprefix = {arXiv},
  primaryclass  = {cs.LG},
  url           = {https://arxiv.org/abs/2011.13456}
}

@misc{chen2025blip3ofamilyfullyopen,
      title={BLIP3-o: A Family of Fully Open Unified Multimodal Models-Architecture, Training and Dataset}, 
      author={Jiuhai Chen and Zhiyang Xu and Xichen Pan and Yushi Hu and Can Qin and Tom Goldstein and Lifu Huang and Tianyi Zhou and Saining Xie and Silvio Savarese and Le Xue and Caiming Xiong and Ran Xu},
      year={2025},
      eprint={2505.09568},
      archivePrefix={arXiv},
      primaryClass={cs.CV},
      url={https://arxiv.org/abs/2505.09568}, 
}

@misc{sun2023journeydbbenchmarkgenerativeimage,
      title={JourneyDB: A Benchmark for Generative Image Understanding}, 
      author={Keqiang Sun and Junting Pan and Yuying Ge and Hao Li and Haodong Duan and Xiaoshi Wu and Renrui Zhang and Aojun Zhou and Zipeng Qin and Yi Wang and Jifeng Dai and Yu Qiao and Limin Wang and Hongsheng Li},
      year={2023},
      eprint={2307.00716},
      archivePrefix={arXiv},
      primaryClass={cs.CV},
      url={https://arxiv.org/abs/2307.00716}, 
}

@misc{hu2024ellaequipdiffusionmodels,
      title={ELLA: Equip Diffusion Models with LLM for Enhanced Semantic Alignment}, 
      author={Xiwei Hu and Rui Wang and Yixiao Fang and Bin Fu and Pei Cheng and Gang Yu},
      year={2024},
      eprint={2403.05135},
      archivePrefix={arXiv},
      primaryClass={cs.CV},
      url={https://arxiv.org/abs/2403.05135}, 
}

@inproceedings{Cherti_2023,
   title={Reproducible Scaling Laws for Contrastive Language-Image Learning},
   url={http://dx.doi.org/10.1109/CVPR52729.2023.00276},
   DOI={10.1109/cvpr52729.2023.00276},
   booktitle={2023 IEEE/CVF Conference on Computer Vision and Pattern Recognition (CVPR)},
   publisher={IEEE},
   author={Cherti, Mehdi and Beaumont, Romain and Wightman, Ross and Wortsman, Mitchell and Ilharco, Gabriel and Gordon, Cade and Schuhmann, Christoph and Schmidt, Ludwig and Jitsev, Jenia},
   year={2023},
   month={June}, pages={2818--2829} }

@misc{ghosh2023genevalobjectfocusedframeworkevaluating,
      title={GenEval: An Object-Focused Framework for Evaluating Text-to-Image Alignment}, 
      author={Dhruba Ghosh and Hanna Hajishirzi and Ludwig Schmidt},
      year={2023},
      eprint={2310.11513},
      archivePrefix={arXiv},
      primaryClass={cs.CV},
      url={https://arxiv.org/abs/2310.11513}, 
}

@misc{ho2022classifierfreediffusionguidance,
      title={Classifier-Free Diffusion Guidance}, 
      author={Jonathan Ho and Tim Salimans},
      year={2022},
      eprint={2207.12598},
      archivePrefix={arXiv},
      primaryClass={cs.LG},
      url={https://arxiv.org/abs/2207.12598}, 
}
